\documentclass{article}

\usepackage{microtype}
\usepackage{graphicx}
\usepackage{subcaption}
\usepackage{booktabs} % for professional tables
\usepackage[table]{xcolor}
\usepackage{siunitx}
\usepackage{ulem} % for \uline (or replace \uline with \underline)
\usepackage{makecell}

\usepackage{hyperref}

\usepackage{caption}  
\usepackage{needspace}

\usepackage[preprint]{icml2026}

\usepackage{amsmath}
\usepackage{amssymb}
\usepackage{mathtools}
\usepackage{amsthm}

\usepackage{enumitem}

\usepackage[capitalize,noabbrev]{cleveref}

\theoremstyle{plain}

\theoremstyle{definition}

\theoremstyle{remark}

\usepackage[textsize=tiny]{todonotes}

\begin{document}

\twocolumn[
\icmltitle{Semantic Routing: Exploring Multi-Layer LLM Feature Weighting for Diffusion Transformers}

%Dynamic Guidance: Exploring Temporal and Depth-Wise Fusion of Multi-Layer LLM Features for Diffusion Transformers

%Semantic Routing: Exploring Time-Wise and Layer-Wise Weighting of Multi-Layer LLM Features for Diffusion Transformers

%Where Meets When: Exploring Layer-Wise and Time-Wise Fusion Strategies for Multi-Layer LLM Conditioning in DiTs




% 现有这行保留
\icmlsetsymbol{intern}{*}
% 新增：给“projector leader”加一个符号（这里用 †）
\icmlsetsymbol{pleader}{$\dagger$}

\begin{icmlauthorlist}
\icmlauthor{Bozhou Li}{pku,kling,intern}
\icmlauthor{Yushuo Guan}{kling,pleader}
\icmlauthor{Haolin Li}{fdu}
\icmlauthor{Bohan Zeng}{pku}
\icmlauthor{Yiyan Ji}{nju}
\icmlauthor{Yue Ding}{cas}
\icmlauthor{Pengfei Wan}{kling}
\icmlauthor{Kun Gai}{kling}
\icmlauthor{Yuanxing Zhang}{kling}
\icmlauthor{Wentao Zhang}{pku}
\end{icmlauthorlist}

\icmlaffiliation{pku}{Peking University}
\icmlaffiliation{kling}{Kling Team, Kuaishou Technology }
\icmlaffiliation{cas}{School of Artificial Intelligence, University of Chinese Academy of Sciences}
\icmlaffiliation{nju}{Nanjing University}
\icmlaffiliation{fdu}{Fudan University}

\vspace{-0.2cm}
\begin{center}
Code: \url{https://github.com/zooblastlbz/SemanticRouting}
\end{center}
\vspace{-0.2cm}

\icmlcorrespondingauthor{Bozhou Li}{libozhou@pku.edu.cn}
\icmlcorrespondingauthor{Wentao Zhang}{wentao.zhang@pku.edu.cn}

%\icmlkeywords{Machine Learning, ICML}

\vskip 0.2in
]

% 原有说明保留，同时新增 † 的说明
\printAffiliationsAndNotice{${}^*$ Work done during an internship at Kling Team, Kuaishou Technology. ${}^\dagger$ Projector leader.}




\begin{abstract}
Recent DiT-based text-to-image models increasingly adopt LLMs as text encoders, yet text conditioning remains largely static and often utilizes only a single LLM layer, despite pronounced semantic hierarchy across LLM layers and non-stationary denoising dynamics over both diffusion time and network depth. To better match the dynamic process of DiT generation and thereby enhance the diffusion model's generative capability, we introduce a unified normalized convex fusion framework equipped with lightweight gates to systematically organize multi-layer LLM hidden states via time-wise, depth-wise, and joint fusion. Experiments establish Depth-wise Semantic Routing as the superior conditioning strategy, consistently improving text–image alignment and compositional generation (e.g., +9.97 on the GenAI-Bench Counting task). Conversely, we find that purely time-wise fusion can paradoxically degrade visual generation fidelity. We attribute this to a train–inference trajectory mismatch: under classifier-free guidance, nominal timesteps fail to track the effective SNR, causing semantically mistimed feature injection during inference. Overall, our results position depth-wise routing as a strong and effective baseline and highlight the critical need for trajectory-aware signals to enable robust time-dependent conditioning. 
\end{abstract}

\section{Introduction}

\begin{figure}[t]
    \centering
    \includegraphics[width=\linewidth]{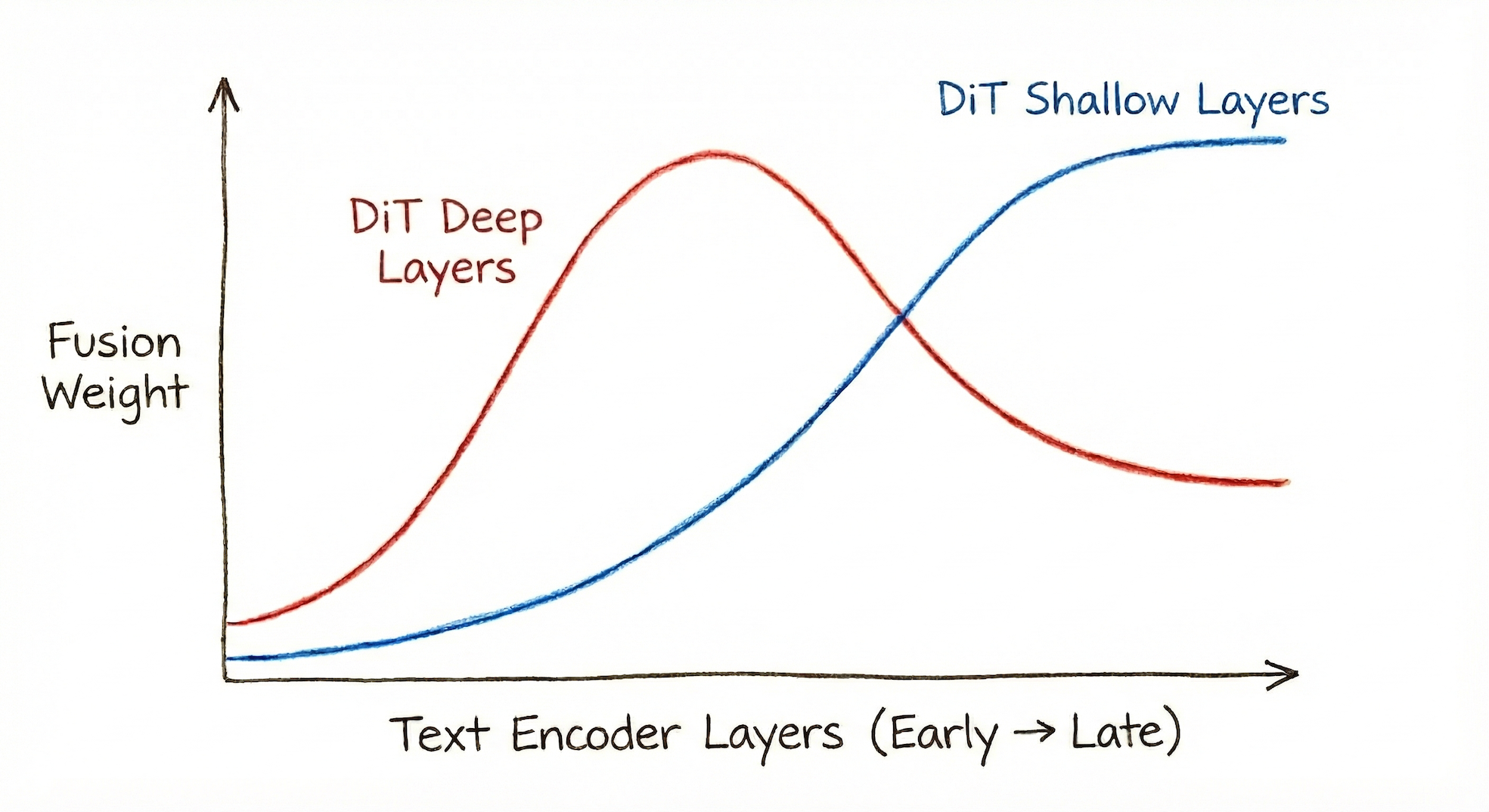}
    % \caption{\textbf{Learned Fusion Weights.} We observe distinct weight distributions for shallow (blue) versus deep (red) DiT blocks, indicating that different generative stages utilize distinct semantic levels from the text encoder.}
    \caption{\textbf{Learned Fusion Weights.} We observe distinct weight distributions for \textcolor[rgb]{0.2078,0.4196,0.6667}{shallow} versus \textcolor[rgb]{0.694,0.227,0.169}{deep} DiT blocks, indicating that different generative stages utilize distinct semantic levels from the text encoder.}
    \label{fig:teaser_stratification}
\end{figure}

%Currently, diffusion models have established dominance in image and video generation \citep{esser2024scaling,gao2025seedance,wu2025qwen,kong2024hunyuanvideo,ma2025step,cai2025z,wan2025wan}. Within prevailing diffusion architectures, particularly Diffusion Transformers (DiTs \citep{peebles2023scalable}), the text encoder is a pivotal component that extracts textual features as conditional inputs to guide the visual synthesis process.

%The text encoders used in diffusion models have witnessed a paradigm shift, moving from encoder-only structures like T5~\citep{raffel2020exploring} and CLIP~\citep{radford2021learning} to decoder-only LLMs for their superior semantic expressiveness~\citep{behnamghader2024llm2vec,ma2024exploring,wang2025comprehensive,li2025gran}. 

Diffusion models have established dominance in image and video generation~\citep{esser2024scaling,gao2025seedance,wu2025qwen,kong2024hunyuanvideo,ma2025step,cai2025z,wan2025wan}, particularly via Diffusion Transformers (DiTs; \citealp{peebles2023scalable}). 
As the pivotal component guiding visual synthesis, the text encoder has recently witnessed a paradigm shift: moving from encoder-only architectures like T5~\citep{raffel2020exploring} and CLIP~\citep{radford2021learning} to decoder-only LLMs to leverage their superior semantic expressiveness~\citep{behnamghader2024llm2vec,ma2024exploring,wang2025comprehensive,li2025gran}.

However, while text encoders have become increasingly potent, the underlying conditioning mechanisms remain static: most prevailing methods inject a fixed text representation, failing to account for the \textbf{evolving conditioning needs across both diffusion time and network depth}. 

This discrepancy becomes particularly salient when considering the multi-faceted heterogeneity in both LLMs and DiTs. 
Figure~\ref{fig:teaser_stratification} illustrates this functional stratification, depicting distinct semantic preferences between shallow and deep DiT blocks. 
From the LLM perspective, representations are inherently hierarchical: layers at varying depths capture distinct semantic granularities and levels of conceptual abstraction~\citep{liu2024fantastic,jin2025exploring,fan2024not,barbero2025llms,skean2025layer}. 
Concurrently, from the DiT perspective, the diffusion trajectory is intrinsically non-uniform across dual axes: (i) temporally, denoising progresses through a coarse-to-fine evolution, where earlier timesteps prioritize low-frequency global structures while later stages focus on high-frequency textural refinement; and (ii) structurally, DiT blocks exhibit functional stratification, contributing unevenly to structural formation versus detail synthesis~\citep{chen2024deltadittrainingfreeaccelerationmethod}.

These observations raise a fundamental research question: To fully unleash the \textbf{immense capacity of billion-parameter LLMs}, how can we \textbf{adaptively route and aggregate} hierarchical signals through a mechanism \textbf{conditioned on diffusion timesteps and distinct DiT network depths}, thereby extracting superior text representations to enhance the overall generative pipeline?

In this work, we take a systematic empirical perspective on text conditioning with multi-layer LLM features. We investigate feature fusion along two orthogonal axes: (i) time-wise adaptivity, where fusion weights vary with the diffusion timestep, and (ii) depth-wise adaptivity, where weights vary with the DiT block index. We further study their combination to understand whether these axes are complementary or introduce undesirable interactions. To enable rigorous, controlled comparisons under matched settings, we develop a unified and lightweight framework that instantiates these alternatives under a single interface with minimal architectural changes, thereby isolating the effects of different fusion strategies.
Our contributions are as follows:
\begin{itemize}[leftmargin=*]
    \item \textbf{Unified Framework for Semantic Routing.} We propose a unified formulation that generalizes text conditioning by dynamically weighting multi-layer LLM features. This lightweight framework instantiates time-wise, depth-wise, and joint gating mechanisms, enabling rigorous, controlled comparisons of adaptive strategies.
    
    \item \textbf{Superiority of Depth-wise Fusion.} \textbf{We establish Depth-wise Semantic Routing as the optimal strategy.} By aligning LLM hierarchy with DiT functional depth, it significantly enhances compositional generation, yielding a substantial \textbf{+9.97} improvement on GenAI-Bench Counting over the penultimate-layer baseline and \textbf{+5.45} over uniform averaging.
    
    \item \textbf{Diagnosis of Trajectory Mismatch.} We identify a critical failure mode in time-wise fusion: a \textbf{train--inference trajectory mismatch}. We demonstrate that nominal timesteps fail to track the effective SNR during iterative sampling, causing semantic misalignment. This mechanistic insight motivates future designs for trajectory-aware conditioning to enable robust time-dependent fusion.
\end{itemize}

\section{Related Work}
\label{sec:related_work}

\subsection{Text Conditioning and Text Encoders }
\label{sec:2.1}
%Early U-Net-based diffusion models primarily relied on cross-attention mechanisms to inject text conditions into the denoising network~\citep{ronneberger2015u}. In the DiT paradigm, \citet{peebles2023scalable} demonstrated that adaLN-Zero can serve as an effective conditioning interface. As adaLN-Zero is inherently tailored for global conditioning, PixArt-$\alpha$ reintroduced cross-attention to enable finer-grained control, a design choice that has become prevalent in subsequent DiT architectures~\citep{chen2023pixart}. Beyond these designs, MMDiT further unifies image and text tokens by applying joint self-attention over their concatenated sequences~\citep{esser2024scaling}.

Text conditioning has evolved from U-Net cross-attention~\citep{ronneberger2015u} to DiT's adaLN-Zero~\citep{peebles2023scalable}. To enable fine-grained control beyond global adaLN, PixArt-$\alpha$ reintroduced cross-attention~\citep{chen2023pixart}, while MMDiT further unified modalities via joint self-attention~\citep{esser2024scaling}.

Regarding text encoders, early diffusion models predominantly employed encoder-only architectures such as CLIP~\citep{radford2021learning} or T5~\citep{raffel2020exploring}. With the rapid evolution of LLMs, recent research has pivoted toward leveraging decoder-only LLMs to obtain stronger semantics for diffusion conditioning~\citep{behnamghader2024llm2vec,ma2024exploring,wang2025comprehensive,li2025gran}. Due to the suboptimal performance of early iterations like LLaMA-2~\citep{touvron2023llama} in this context, LiDiT introduced a refiner module to enhance extracted text features~\citep{ma2024exploring}. However, with the advent of more capable foundation models, the reliance on such auxiliary components has diminished~\citep{seedream2025seedream,wu2025qwen,cai2025z,gao2025seedance,kong2024hunyuanvideo,ma2025step}. Consequently, the field has gravitated toward direct utilization of LLM representations, where standard practice typically adopts single-layer features, predominantly from the penultimate layer.

\subsection{Semantic Heterogeneity and Multi-Layer Fusion}
\label{sec:2.2}
LLMs exhibit distinct representational characteristics across their hierarchy. Probing techniques~\citep{liu2024fantastic} have elucidated that shallow layers primarily capture lexical semantics, whereas deeper layers are increasingly shaped by next-token prediction objectives. Complementarily, \citet{jin2025exploring} observed that complex conceptual abstractions are typically acquired in deeper layers. Regarding task specificity, the utility of different layers varies across downstream applications~\citep{fan2024not}, while \citet{gurnee2023language} revealed that spatial and temporal information can be encoded in distinct layers. Notably, intermediate layers offer unique advantages, such as attenuated attention sinks~\citep{barbero2025llms} and high semantic compression~\citep{skean2025layer}, highlighting the potential of leveraging multi-layer semantics to enhance text conditioning.

%This hierarchical heterogeneity suggests that combining information across LLM layers has the potential to provide more expressive textual conditioning for diffusion models; in parallel, recent empirical evidence has demonstrated that aggregating multi-layer features can yield superior performance over single-layer counterparts~\citep{wang2025comprehensive}.

Such heterogeneity highlights the potential of multi-layer conditioning, while recent evidence confirms its superiority over single-layer baselines~\citep{wang2025comprehensive}.  For instance, Playground v3~\citep{liu2024playground} and \citet{tang2025exploring} adopt a deep fusion strategy where internal attention K/V states from the LLM are directly reused in DiT's cross-attention, in pursuit of a deeper  fusion between the text encoder and the DiT backbone. However, the former lacks rigorous controlled experiments to isolate the source of its improvements, while the latter primarily evaluates its approach against the final LLM layer  rather than the typically used penultimate layer. While normalizing multi-layer features~\citep{wang2025comprehensive} or introducing learnable weights for adaptive fusion~\citep{li2025gran} has shown promise, these conditioning mechanisms typically remain static, applying a uniform fusion strategy regardless of the diffusion timestep or the DiT block index.

\subsection{Temporal and Depth-wise Dynamics in DiT}
\label{sec:2.3}

The diffusion generative process is characterized by distinct, non-uniform dynamics along both temporal and structural dimensions. From a temporal perspective, the denoising trajectory follows a coarse-to-fine paradigm: earlier timesteps prioritize the recovery of low-frequency global structures, whereas later stages transition toward the refinement of high-frequency textures and details~\citep{hertz2022prompt,liu2023oms,wang2023diffusion}. This temporal non-stationarity suggests that the demand for textual guidance may evolve across different denoising stages. From a depth-wise perspective, DiTs exhibit functional stratification along the network hierarchy: shallower blocks are primarily responsible for structural formation, while deeper blocks contribute more to detail synthesis~\citep{chen2024deltadittrainingfreeaccelerationmethod}. These  variations across time and depth prompt us to consider whether multi-layer LLM semantics can be effectively fused via time-wise adaptivity, depth-wise adaptivity, or their combination, aimed at enhancing generative models from the perspective of text conditioning.
\section{Method}
\label{sec:method}

\subsection{Problem Setup}
\label{sec:method_overview}
This work investigates text conditioning mechanisms within DiTs under the flow matching formulation \citep{lipman2022flow}. Specifically, the conditioning signal is synthesized by aggregating hidden-state sequences across multiple layers of a pretrained LLM. Building upon the observations in Section~\ref{sec:related_work}, we identify two primary sources of variation that govern the utility of different semantic abstractions: the flow time $t \in [0,1]$ and the DiT block index $d \in \{1,\dots,D\}$ (representing network depth). Consequently, we systematically evaluate fusion mechanisms where weights are conditioned on $t$ (time-wise), $d$ (depth-wise), or both (jointly). To ensure a controlled comparison, all variants employ an identical DiT backbone and training protocol, differing exclusively in the design of the fusion module.

\subsection{Preliminaries and Notation}
\label{sec:notation}

\paragraph{Flow matching and timestep.}
Let $x(t)$ denote the sample state (or latent) at continuous time $t \in [0,1]$ along the flow, with marginal distribution $x(t) \sim p_t$, where $p_0$ is a simple base distribution (e.g., Gaussian noise) and $p_1$ is the target data distribution. A DiT-based backbone parameterizes a text-conditioned vector field
 $v_{\theta}\big(x(t), t, c\big)$, where $c$ denotes the text condition. Given $c$ and an initial condition $x(0) \sim p_0$, sampling is performed by integrating the ODE
\begin{equation}
\frac{d x(t)}{d t} = v_{\theta}\big(x(t), t, c\big),
\end{equation}
which transports the sample from $p_0$ toward $p_1$. For consistency, we refer to $t$ as the timestep throughout this paper.

\paragraph{DiT depth and conditioning site.}
The DiT backbone consists of $D$ Transformer blocks. We use $d \in \{1,\dots,D\}$ to index the specific block where the conditioning signal is applied. In our experimental setup, the fused text representation $H_{\text{cond}}(t,d)$ provides the conditioning sequence of text hidden states that is fed to the cross-attention module in block $d$.

\paragraph{Multi-layer LLM features.}
Let the pretrained LLM provide hidden-state sequences from its entire hierarchy. Let $\mathcal{L}$ denote this set of layers, with $|\mathcal{L}|=L$. We represent the sequence output from layer $l \in \mathcal{L}$ as $H^{(l)} \in \mathbb{R}^{N \times C}$, where $N$ is the text sequence length and $C$ is the LLM hidden dimension.

\subsection{A Unified Formulation for Multi-layer Fusion}
\label{sec:fusion_formulation}

The hierarchical nature of LLM representations facilitates the capture of complementary linguistic information across a continuum of abstraction levels. Concurrently, the conditioning demands of a generative model are inherently non-stationary: they evolve both across the flow time $t$  and across DiT depth $d$ due to the functional stratification of Transformer blocks. To investigate and mitigate this potential misalignment between semantic supply and conditioning demand, we propose a unified formulation that parameterizes the interaction between the LLM hierarchy and DiT dynamics. This framework enables flexible semantic routing across both temporal and structural axes, subsuming the diverse set of fusion strategies investigated in this study as specific instances.

We instantiate the text condition using a normalized convex fusion of multi-layer features. Specifically, we apply LayerNorm ~\citep{ba2016layer} to each layer-wise feature to mitigate scale discrepancies across the hierarchy \citep{kim2025peri,li2025unseen}. The final fused representation is formed via a softmax-normalized convex combination, which ensures the resulting feature remains within the convex hull of the normalized layer representations, rendering the learned weights directly interpretable:
\begin{equation}
H_{\text{cond}}(t,d) = \sum_{l \in \mathcal{L}} \alpha^{(l)}_{t,d} \cdot \mathrm{LN}\!\left(H^{(l)}\right),
\label{eq:fusion_general}
\end{equation}
where the weights $\alpha_{t,d} = \{\alpha^{(l)}_{t,d}\}_{l \in \mathcal{L}} \in \mathbb{R}^{L}$ satisfy $\sum_{l} \alpha^{(l)}_{t,d} = 1$ and $\alpha^{(l)}_{t,d} \ge 0$. These weights are derived by applying a softmax function to the logits $z_{t,d} \in \mathbb{R}^{L}$:
\begin{equation}
\alpha_{t,d} = \mathrm{Softmax}\!\left(z_{t,d}\right).
\label{eq:alpha_softmax}
\end{equation}
Different fusion strategies correspond to distinct parameterizations of the logits $z_{t,d}$.

\subsection{Fusion Weight Parameterizations}
\label{sec:fusion_strategies}
We evaluate the following parameterization schemes under the framework of Eq.~\eqref{eq:fusion_general}.

\paragraph{(B1) Penultimate-layer baseline.}
We utilize only the penultimate LLM layer for conditioning:
\begin{equation}
H_{\text{cond}}(t,d) = \mathrm{LN}\!\left(H^{(l^\star)}\right), \quad l^\star = \text{penultimate}.
\label{eq:penultimate}
\end{equation}

\paragraph{(B2) Uniform normalized averaging.}
We aggregate all layers via a uniform average after normalization, without introducing learnable fusion weights:
\begin{equation}
H_{\text{cond}}(t,d) = \frac{1}{L} \sum_{l \in \mathcal{L}} \mathrm{LN}\!\left(H^{(l)}\right).
\label{eq:uniform_norm_avg}
\end{equation}

\paragraph{(B3) Static learnable fusion.}
We learn a single global logit vector shared across all pairs of $(t,d)$:
\begin{equation}
z_{t,d} = \beta, \quad \beta \in \mathbb{R}^{L} \text{ (learnable)}.
\label{eq:static_learnable}
\end{equation}

\paragraph{Time-conditioned fusion gate (TCFG).}
To facilitate time-dependent adaptivity, we introduce a lightweight gating module that maps the flow time $t$ to fusion logits. We first embed the continuous time using a sinusoidal encoding $\phi(t)$ and subsequently compute the logits via a small MLP:
\begin{equation}
z_{t} = g_{\psi}\big(\phi(t)\big), \quad g_{\psi} = \mathrm{MLP}(\cdot),
\label{eq:tcfg}
\end{equation}
where $z_t \in \mathbb{R}^{L}$ yields fusion weights via Eq.~\eqref{eq:alpha_softmax}. The TCFG serves as the fundamental building block for both time-wise and joint fusion strategies. More details can be seen in Appendix ~\ref{app:TCFG_details}.

\paragraph{(S1) Time-wise fusion.}
We apply a shared TCFG across all DiT blocks, making the fusion dependent on $t$ but invariant to $d$:
\begin{equation}
z_{t,d} = z_{t} = g_{\psi}\big(\phi(t)\big).
\label{eq:timewise}
\end{equation}

\paragraph{(S2) Depth-wise fusion.}
We learn block-specific logits that depend on the depth index $d$ but remain static over time $t$:
\begin{equation}
z_{t,d} = z_{d} = \beta_{d}, \quad \beta_{d} \in \mathbb{R}^{L} \text{ (learnable for each } d).
\label{eq:depthwise}
\end{equation}

\paragraph{(S3) Joint time-and-depth fusion.}
We model the dependency on both $t$ and $d$ by employing a depth-specific TCFG for each DiT block. Concretely, each block $d$ has its own gating function $g_{\psi_d}$:
\begin{equation}
z_{t,d} = g_{\psi_d}\big(\phi(t)\big).
\label{eq:joint_perblock_gate}
\end{equation}
The weights $\alpha_{t,d}$ are then obtained via Eq.~\eqref{eq:alpha_softmax} to compute $H_{\text{cond}}(t,d)$ via Eq.~\eqref{eq:fusion_general}.

\section{Experiments}
\label{sec:exp}

\subsection{Experimental Setup}
\label{sec:exp_setup}

\paragraph{Models.}
We employ Qwen3-VL-4B-Instruct \citep{Qwen3-VL} as the text encoder and the pretrained VAE from Stable Diffusion 3 (SD3) \citep{esser2024scaling}. The diffusion backbone is a cross-attention-based DiT comprising $D=24$ Transformer blocks and approximately 2.24B parameters. The architectural design of the backbone follows the implementation of FuseDiT ~\citep{tang2025exploring}: we apply 1D RoPE ~\citep{su2024roformer} to text prompts, 2D RoPE ~\citep{heo2024rotary} to image latents, and use QK-Norm ~\citep{henry2020query} in attention. The only architectural difference from FuseDiT is that we do not use Sandwich Norm ~\citep{gong2022sandwich}. Unless otherwise specified, all conditioning variants in Section~\ref{sec:method} share this identical backbone and differ only in how multi-layer text features are fused. For the TCFG module, we utilize a 128-dimensional sinusoidal encoding for the timestep input.

\paragraph{Dataset.}
We train all models on a high-quality subset of LAION-400M ~\citep{schuhmann2021laion}, comprising approximately 30 million image-text pairs. We replace the original  texts with dense synthetic captions generated by Qwen3-VL-32B-Instruct ~\citep{Qwen3-VL}. Images are resized to $256 \times 256$, and text prompts are tokenized with a maximum length of 512 tokens.

\paragraph{Training.}
We train all models using AdamW ~\citep{loshchilov2017decoupled} with $(\beta_1,\beta_2)=(0.9,0.999)$, a learning rate of $1\times10^{-4}$, weight decay of $1\times10^{-4}$, and a constant learning rate scheduler.
The batch size is  512, and all models are trained for 500k steps.
The prompt drop ratio is set to 0.1 to enable unconditional generation.
For timestep sampling, we follow the logit-normal distribution used in  SD3.

\paragraph{Baselines.}
We evaluate our proposed strategies against two categories of baselines: 
(i) the \textbf{Standard Baselines} (B1--B3) introduced in Section~\ref{sec:fusion_strategies}; 
and (ii) \textbf{FuseDiT} \citep{tang2025exploring}, a representative deep-fusion approach that reuses internal LLM attention K/V states directly within the DiT attention layers to facilitate a more intrinsic integration between the text encoder and the visual backbone. 

\paragraph{Evaluation.}
Unless otherwise specified, images are generated using 50 sampling steps with a CFG ~\citep{ho2022classifier} scale of $6.0$ and the FlowMatch Euler scheduler from diffusers ~\citep{von-platen-etal-2022-diffusers}.
We use GenAI-Bench ~\citep{li2024genai} and GenEval ~\citep{ghosh2023geneval} to assess text--image alignment in generated samples. For GenAI-Bench, evaluation is conducted using Qwen3-VL-235B-A22B-Instruct as the judge model. To quantify aesthetic appeal, we report the style dimension scores from UnifiedReward-2.0 ~\citep{unifiedreward} on samples generated using the DrawBench prompt set ~\citep{saharia2022photorealistic}.

\subsection{Main Results}
\label{sec:main_results}

\begin{table}[ht]
  \centering
  \caption{Performance of different fusion strategies on three benchmarks. Best in each column is in \textbf{bold}, and second best is \uline{underlined}.}
  \label{tab:main_results}
  \setlength{\tabcolsep}{5pt}
  \small
  \begin{tabular}{l
                  S[table-format=2.2]
                  S[table-format=2.2]
                  S[table-format=1.2]}
    \toprule
    Method & {GenEval $\uparrow$} & {GenAI $\uparrow$} & {UnifiedReward $\uparrow$} \\
    \midrule

    \multicolumn{4}{l}{\textit{Baselines}} \\
    B1: Penult.  & 64.54 & 74.96 & 3.02 \\
    B2: Uniform  & \uline{66.51} & 76.82 & \textbf{3.06} \\
    B3: Static   & 64.77 & 76.31 &
    \uline{3.05} \\
    \midrule

    \multicolumn{4}{l}{\textit{Deep-fusion baseline}} \\
    FuseDiT      & 60.95 & 75.02 & \uline{3.05} \\
    \midrule

    \multicolumn{4}{l}{\textit{Our fusion strategies}} \\
    S1: Time     & 63.41 & 76.20 & 2.97 \\
    S2: Depth    & \textbf{67.07} & \textbf{79.07} & \textbf{3.06} \\
    S3: Joint    & 66.05 & \uline{77.44} & \textbf{3.06} \\
    \bottomrule
  \end{tabular}
\end{table}

We report the overall performance in Table~\ref{tab:main_results} and the granular capability breakdown on GenAI-Bench in Table~\ref{tab:genai_subtasks}. The empirical results summarized in these tables reveal several key insights:

\paragraph{Limits of Static Aggregation.}
First, aggregating multi-layer features (B2--S3) consistently outperforms the penultimate-layer baseline (B1). This trend suggests that LLM hierarchies contain complementary semantic signals that are largely underutilized by conventional single-layer conditioning. Furthermore, the learnable static fusion (B3) fails to surpass uniform normalized averaging (B2), indicating that without explicit adaptivity, a fixed set of learned weights is not robust enough to outperform a strong uniform prior.

\paragraph{Interplay in Deep Fusion Architectures.}
Regarding the deep-fusion baseline, FuseDiT, Table~\ref{tab:main_results} reveals that its architecture struggles to effectively extract essential textual information for high-quality synthesis. We hypothesize that this performance gap stems from an inherent architectural constraint: by directly reusing internal LLM key/value states within cross-attention, FuseDiT imposes a restrictive coupling that deprives the DiT backbone of the flexibility to dynamically re-contextualize text features. These observations offer a useful perspective for future unified architectures, highlighting the importance of carefully considering the interplay between internal state-sharing mechanisms and the task-specific feature extraction capabilities required for high-fidelity generative modeling.

\paragraph{Superiority of Depth-wise Semantic Routing.}
Among our proposed strategies, \textbf{depth-wise fusion (S2)} delivers the most robust and significant overall gains. Notably, introducing learnable weights in S2 yields clear improvements over B2 (which can be viewed as a fixed-weight depth-wise scheme). This contrast with the static setting (where learnable B3 fails to surpass B2) highlights a critical divergence: purely global parameterization is ineffective; rather, the value of learnability is effectively unlocked only when aligned with the depth-wise structural hierarchy.

This advantage implies that hierarchical LLM semantics are indispensable for navigating intricate prompts. As shown in Table~\ref{tab:genai_subtasks}, performance gains are disproportionately pronounced within ``Advanced'' categories. Taking the ``Counting'' task as a representative case, S2 achieves a substantial improvement of \textbf{+9.97} over B1 and \textbf{+5.45} over B2. This disparity underscores a pivotal insight: naive aggregation of multiple layers is insufficient to unlock the full potential of text conditioning. Rather, the synergy created by allowing functional blocks at different DiT depths to selectively route and aggregate task-relevant LLM-layer semantics is the key to mastering compositional reasoning and fine-grained constraint following.

\paragraph{Instability of Time-Awareness and Joint Mitigation.}
In contrast, purely time-wise fusion (S1) does not provide consistent benefits and often leads to degraded generation quality, manifesting as noticeable blurriness and loss of fine details (see Appendix~\ref{app:img}). We provide a detailed mechanistic diagnosis of this phenomenon in Section~\ref{subsec:misalignment}, attributing it to optimization conflicts arising from fundamental train--inference inconsistencies along the temporal axis. Joint fusion (S3) remains competitive but is slightly less effective than S2. Notably, S3 avoids the blurriness characteristic of S1 by incorporating depth-specificity, a phenomenon further analyzed in Section~\ref{subsec:misalignment}.

\setlength{\abovecaptionskip}{7.2pt}
\setlength{\belowcaptionskip}{7.2pt}

% colored signed changes (relative to B1): green = increase, red = decrease
\newcommand{\dpos}[1]{\textcolor{green!50!black}{#1}}
\newcommand{\dneg}[1]{\textcolor{red!70!black}{#1}}
\newcommand{\dzero}{\textcolor{gray}{+0.00}}

% two-line cells: score + colored signed change
\newcommand{\celld}[2]{\makecell[c]{#1\\[-1pt]{\scriptsize #2}}}
\newcommand{\celldb}[2]{\makecell[c]{\textbf{#1}\\[-1pt]{\scriptsize\textbf{#2}}}}
\newcommand{\celldu}[2]{\makecell[c]{\uline{#1}\\[-1pt]{\scriptsize\uline{#2}}}}
% ---------- End Preamble ----------

\begin{table*}[!htb]
  \centering
\caption{Fine-grained GenAI-Bench performance.
\textbf{Basic skills} include Attribute, Scene, Spatial relations, Action relations, and Part relations.
\textbf{Advanced skills} include Counting, Comparison, Differentiation, Negation, and Universal.
We report average scores for each group; signed changes relative to B1 are colored (\dpos{green}/\dneg{red}). 
\textbf{Bold} and \uline{underlined} denote best and second-best results, respectively.}
  \label{tab:genai_subtasks}
  \setlength{\tabcolsep}{4.2pt}
  \begin{tabular}{lccc ccccc ccccc}
    \toprule
    & \multicolumn{3}{c}{Summary} & \multicolumn{5}{c}{Basic} & \multicolumn{5}{c}{Advanced} \\
    \cmidrule(lr){2-4}\cmidrule(lr){5-9}\cmidrule(lr){10-14}
    Method & Avg & Basic & Adv. & Attr. & Scene & Spat. & Action & Part & Count. & Comp. & Differ. & Neg. & Uni. \\
    \midrule

    \multicolumn{14}{l}{\textit{Baselines}} \\
    B1: Penult. &
    \celld{74.96}{\dzero} & \celld{80.05}{\dzero} & \celld{70.55}{\dzero} &
    \celld{79.04}{\dzero} & \celld{85.38}{\dzero} & \celld{81.27}{\dzero} & \celld{83.28}{\dzero} & \celld{73.03}{\dzero} &
    \celld{64.60}{\dzero} & \celld{66.70}{\dzero} & \celld{65.39}{\dzero} & \celld{71.21}{\dzero} & \celld{72.76}{\dzero} \\
    B2: Uniform &
    \celld{76.82}{\dpos{+1.86}} & \celld{81.28}{\dpos{+1.23}} & \celld{72.95}{\dpos{+2.40}} &
    \celld{81.10}{\dpos{+2.06}} & \celld{86.07}{\dpos{+0.69}} & \celld{81.98}{\dpos{+0.71}} & \celld{85.01}{\dpos{+1.73}} & \celld{73.13}{\dpos{+0.10}} &
    \celldu{69.12}{\dpos{+4.52}} & \celld{68.19}{\dpos{+1.49}} & \celld{70.49}{\dpos{+5.10}} & \celld{71.51}{\dpos{+0.30}} & \celld{71.10}{\dneg{-1.66}} \\
    B3: Static &
    \celld{76.31}{\dpos{+1.35}} & \celld{80.36}{\dpos{+0.31}} & \celld{72.88}{\dpos{+2.33}} &
    \celld{79.41}{\dpos{+0.37}} & \celld{84.93}{\dneg{-0.45}} & \celld{80.59}{\dneg{-0.68}} & \celld{84.13}{\dpos{+0.85}} & \celldu{75.78}{\dpos{+2.75}} &
    \celld{68.32}{\dpos{+3.72}} & \celld{70.52}{\dpos{+3.82}} & \celld{70.57}{\dpos{+5.18}} & \celldu{74.73}{\dpos{+3.52}} & \celld{68.21}{\dneg{-4.55}} \\
    \midrule

    \multicolumn{14}{l}{\textit{Deep-fusion baseline}} \\
    FuseDiT &
    \celld{75.02}{\dpos{+0.06}} & \celld{77.65}{\dneg{-2.40}} & \celld{72.65}{\dpos{+2.10}} &
    \celld{77.45}{\dneg{-1.59}} & \celld{83.49}{\dneg{-1.89}} & \celld{78.69}{\dneg{-2.58}} & \celld{81.39}{\dneg{-1.89}} & \celld{70.52}{\dneg{-2.51}} &
    \celld{67.23}{\dpos{+2.63}} & \celld{66.15}{\dneg{-0.55}} & \celld{68.48}{\dpos{+3.09}} & \celld{74.16}{\dpos{+2.95}} & \celldu{73.31}{\dpos{+0.55}} \\
    \midrule

    \multicolumn{14}{l}{\textit{Our fusion strategies}} \\
    S1: Time &
    \celld{76.20}{\dpos{+1.24}} & \celld{79.69}{\dneg{-0.36}} & \celldu{73.16}{\dpos{+2.61}} &
    \celld{79.17}{\dpos{+0.13}} & \celld{83.50}{\dneg{-1.88}} & \celld{81.03}{\dneg{-0.24}} & \celld{83.18}{\dneg{-0.10}} & \celld{72.38}{\dneg{-0.65}} &
    \celld{66.37}{\dpos{+1.77}} & \celldu{71.74}{\dpos{+5.04}} & \celldu{71.49}{\dpos{+6.10}} & \celldb{75.79}{\dpos{+4.58}} & \celld{70.98}{\dneg{-1.78}} \\
    S2: Depth &
    \celldb{79.07}{\dpos{+4.11}} & \celldu{82.68}{\dpos{+2.63}} & \celldb{76.03}{\dpos{+5.48}} &
    \celldu{81.67}{\dpos{+2.63}} & \celldb{88.33}{\dpos{+2.95}} & \celldu{83.08}{\dpos{+1.81}} & \celldu{86.10}{\dpos{+2.82}} & \celldb{77.89}{\dpos{+4.86}} &
    \celldb{74.57}{\dpos{+9.97}} & \celldb{72.29}{\dpos{+5.59}} & \celldb{74.31}{\dpos{+8.92}} & \celld{74.05}{\dpos{+2.84}} & \celldb{76.09}{\dpos{+3.33}} \\
    S3: Joint &
    \celldu{77.44}{\dpos{+2.48}} & \celldb{82.92}{\dpos{+2.87}} & \celld{72.71}{\dpos{+2.16}} &
    \celldb{82.16}{\dpos{+3.12}} & \celldu{87.90}{\dpos{+2.52}} & \celldb{85.13}{\dpos{+3.86}} & \celldb{87.08}{\dpos{+3.80}} & \celld{74.16}{\dpos{+1.13}} &
    \celld{67.55}{\dpos{+2.95}} & \celld{69.44}{\dpos{+2.74}} & \celld{70.79}{\dpos{+5.40}} & \celld{72.54}{\dpos{+1.33}} & \celld{72.20}{\dneg{-0.56}} \\
    \bottomrule
  \end{tabular}
\end{table*}

\section{Analysis}
\label{sec:analysis}
\begin{figure*}[!htb]
  \centering

  % ---------- Row 1: one wide image across full page width ----------
  \begin{subfigure}[t]{\textwidth}
    \centering
    \includegraphics[width=\linewidth]{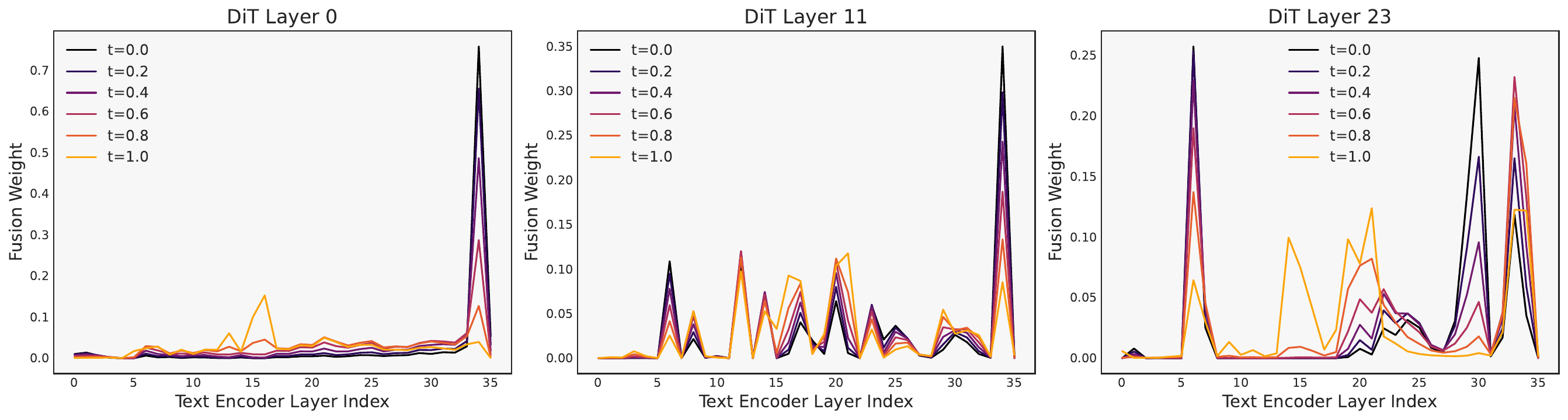}
    \caption{Joint Fusion Weights }
    \label{fig:row1_long}
  \end{subfigure}

  % ---------- Row 2: three narrower images in one row ----------
  \begin{subfigure}[t]{0.33\textwidth}
    \centering
    \includegraphics[width=\linewidth]{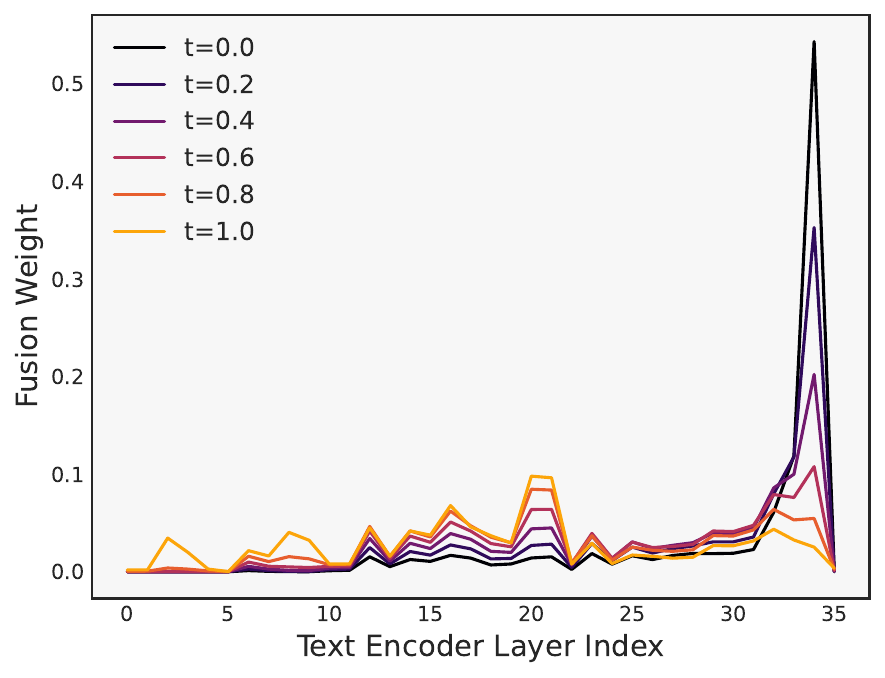}
    \caption{Time-wise Fusion Weights}
    \label{fig:row2_a}
  \end{subfigure}\hfill
  \begin{subfigure}[t]{0.33\textwidth}
    \centering
    \includegraphics[width=\linewidth]{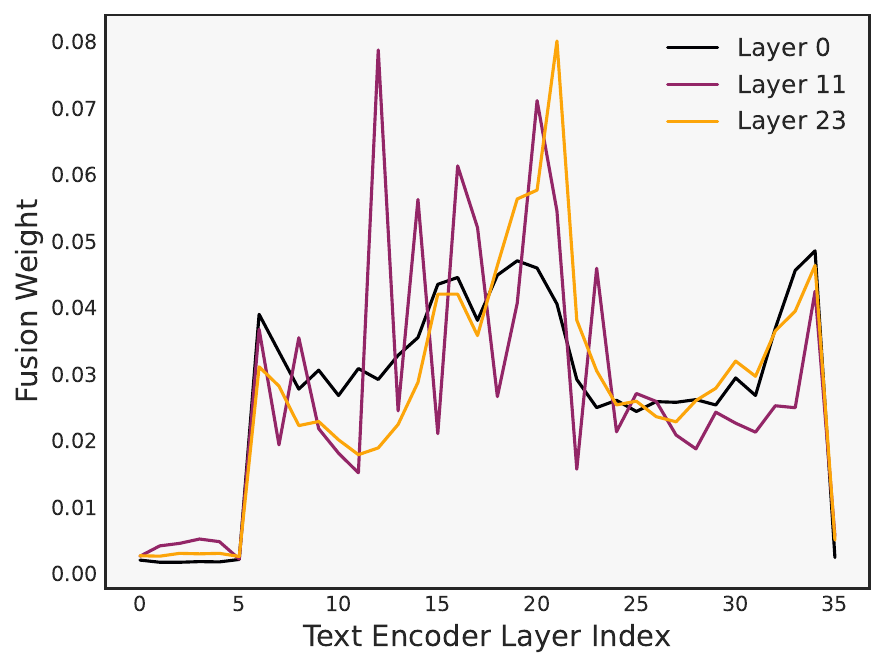}
    \caption{Depth-wise Fusion Weights}
    \label{fig:row2_b}
  \end{subfigure}\hfill
  \begin{subfigure}[t]{0.33\textwidth}
    \centering
    \includegraphics[width=\linewidth]{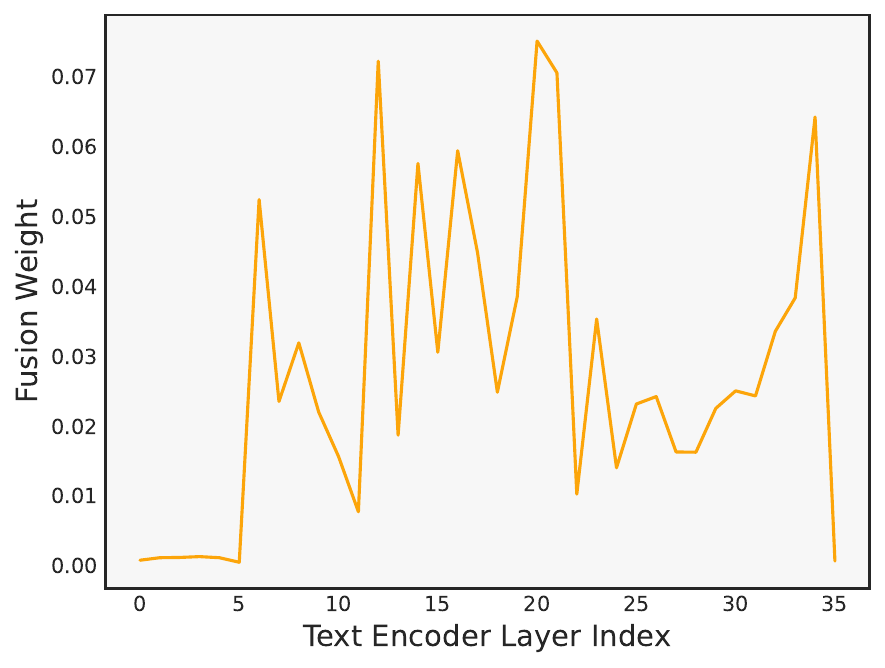}
    \caption{Static Fusion Weights}
    \label{fig:row2_c}
  \end{subfigure}

  \caption{Weight distributions under different fusion-weight parameterizations.
The x-axis denotes the text encoder layer index $l$, and the y-axis denotes the normalized fusion weight $\alpha_{t,d}$.
For time-wise fusion, we sample $t \in [0,1]$ with a step size of $0.2$.
For joint fusion and depth-wise fusion, we report representative DiT block indices $d \in \{0, 11, 23\}$.
Additional visualizations are provided in the appendix ~\ref{app:detail_weights}}.
  \label{fig:two_rows_layout}
\end{figure*}

We analyze the proposed fusion strategies from three aspects: (i)the distribution patterns of fusion weights across flow time $t$ and DiT depth $d$, validating the interpretability of the learned semantic routing; (ii) a mechanistic diagnosis of the degradation observed in purely time-wise fusion, examining how the mismatch between training-time temporal representations and the inference denoising trajectory leads to semantic misalignment; and (iii) the computational overhead introduced by the additional fusion modules.

\subsection{Weight Dynamics over Time and Depth}
\label{sec:analysis_weight_dynamics}

% Preamble:
% \usepackage{graphicx}
% \usepackage{subcaption}

\begin{figure}[!htb]
    \centering
    
    \begin{subfigure}[b]{0.45\linewidth}
        \centering
        \includegraphics[width=\linewidth]{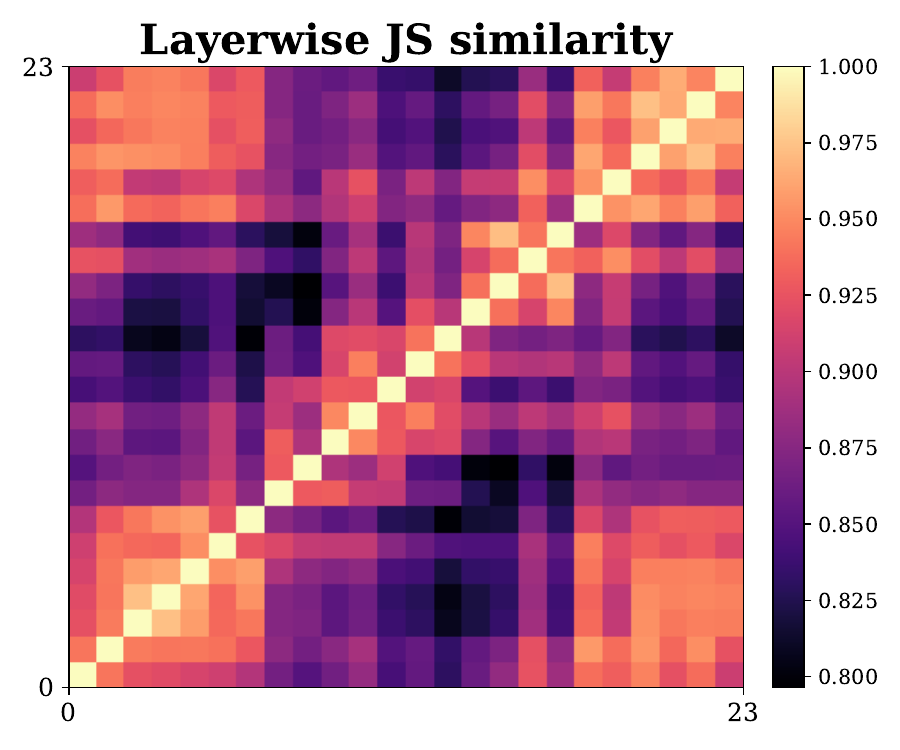} 
        \caption{JS similarity matrix of fusion-weight distributions across DiT blocks.}
        \label{fig:analysis_snr}
    \end{subfigure}
    \hfill
    % 第二幅图
    \begin{subfigure}[b]{0.45\linewidth}
        \centering
        \includegraphics[width=\linewidth]{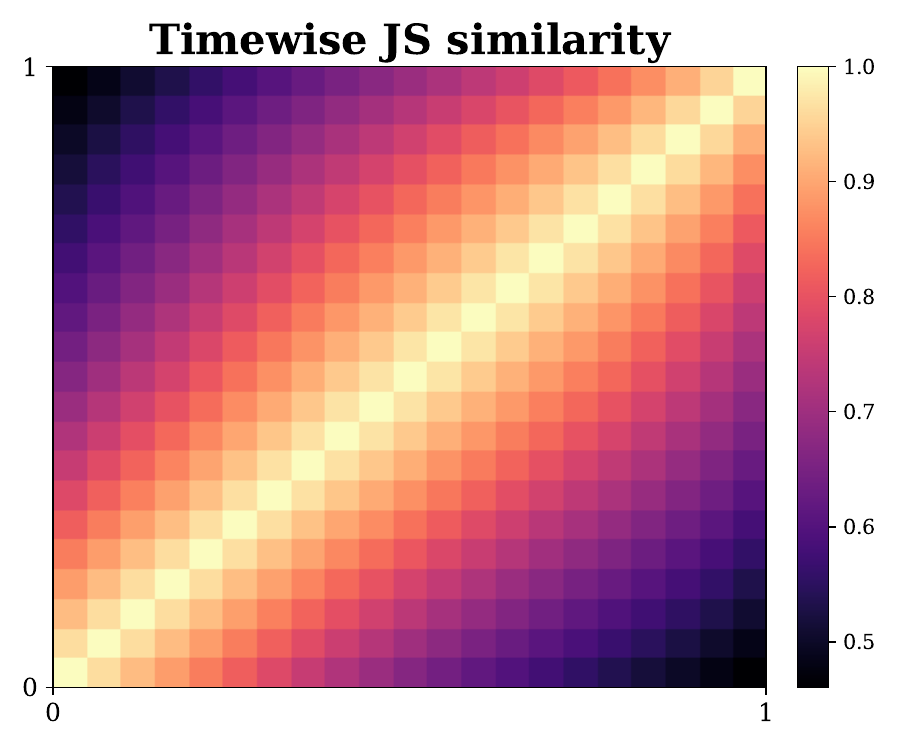}
        \caption{JS similarity matrix of fusion-weight distributions across diffusion timesteps.}
        \label{fig:analysis_fft}
    \end{subfigure}
    
    \caption{Visualizing the local smoothness of learned fusion weights. We compute pairwise JS similarity between normalized fusion-weight distributions along two axes: (a) across DiT blocks (depth) and (b) across diffusion timesteps (time).}
    \label{fig:analysis_simi}
\end{figure}

To validate that the learned fusion weights capture meaningful semantic preferences rather than arbitrary noise, we analyze their evolution across timestep $t$ and DiT depth $d$  in Figure~\ref{fig:two_rows_layout}. More details are provided in Figure~\ref{app:trends_analysis}.

\paragraph{Text Encoder Layer Specificity.}
According to Figure~\ref{fig:two_rows_layout}, the learned weights exhibit clear selectivity: the initial and final text encoder layers consistently receive negligible attention, confirming that effective semantics reside within the model's internal depth. Notably, the penultimate text encoder layer dominates primarily during early timesteps  but fades as generation progresses, as shown in Figure~\ref{fig:two_rows_layout} (a-b). This suggests it serves as a high-level semantic anchor for initial structural layout, while lacking the fine-grained features required for later-stage texture refinement.

\paragraph{Neighbor Inhibition and Information Selection.}
For the intermediate layers of text encoder, as visualized in Figure~\ref{fig:two_rows_layout},  we observe an interesting pattern of weight fluctuation across the layer index. This is characterized by local peaks where specific layers receive high weights while their immediate neighbors are noticeably suppressed. We attribute this to an implicit redundancy reduction strategy learned by the model. Since LLM hidden states typically possess high inter-layer similarity due to residual connections, the gating mechanism tends to select the most representative layer within a local neighborhood to avoid redundant information infusion. This selective inhibition is particularly pronounced in the Joint strategy (Figure~\ref{fig:row1_long}), which exhibits much sharper peaks and higher local contrast compared to the smoother distributions observed in the decoupled Time-wise and Depth-wise settings.

\paragraph{Spatiotemporal Dynamism.}
As shown in Figure~\ref{fig:row2_a}, the learned weights shift significantly across timesteps, reflecting the evolving semantic demands of the denoising process. 
Notably, a comparison between the depth-only setting (Figure~\ref{fig:row1_long}) and the joint setting (Figure~\ref{fig:row2_b}) reveals that S3 exhibits a stronger depth-dependent reallocation. 
This suggests that coupling time and depth enables a more nuanced orchestration of feature selection beyond what is possible with independent dimensions.

\paragraph{Emergent Local Smoothness.}
Similarity visualizations (Figure~\ref{fig:analysis_simi}) confirm that these variations are highly structured. We observe clear smoothness across neighboring timesteps and adjacent DiT blocks.
In the depth-only setting, where no explicit cross-block constraints are enforced, this emergent locality provides robust evidence that the learned routing is driven by coherent semantic signals rather than stochastic optimization noise.

%Our fusion strategies rely on learned weights to combine multi-layer LLM features.
%To validate that the learned weighting is meaningful (rather than arbitrary or degenerate) and to provide interpretability for the observed performance differences, we analyze how these weights evolve across the flow time $t$ and the DiT block depth $d$. Specifically, we analyze the shape of the weight distribution and its variation along the time and depth axes, focusing on whether the allocation is non-uniform and discriminative across LLM layers. As shown in Figure ~\ref{fig:analysis_dynamics}, the weights assigned to different LLM layers become clearly differentiated across both timesteps and DiT block depths, with more pronounced variation over timesteps than over depth. Moreover, compared to the depth-only setting, the joint setting exhibits stronger depth-dependent variation, manifested as a more evident depth-conditioned reallocation of layer weights.

%Furthermore, under the time-wise and depth-wise settings, we visualize similarity matrices of the fusion weights across timesteps and across DiT block depths, respectively (Figure 2). We observe clear local similarity between neighboring timesteps or neighboring blocks, suggesting that the weights vary in a locally smooth manner rather than exhibiting unstructured fluctuations. In the depth-only setting, where there is no explicit cross-block weight sharing or structural constraint, the presence of such locality provides stronger evidence that the weight evolution is not driven by arbitrary noise.

\subsection{Trajectory Misalignment in Time-wise Fusion}
\label{subsec:misalignment}

\begin{figure}[!htb]
    \centering
    
    % --- 第三幅图 (下方) ---
    \begin{subfigure}[b]{0.45\linewidth}
        \centering
        \includegraphics[width=\linewidth]{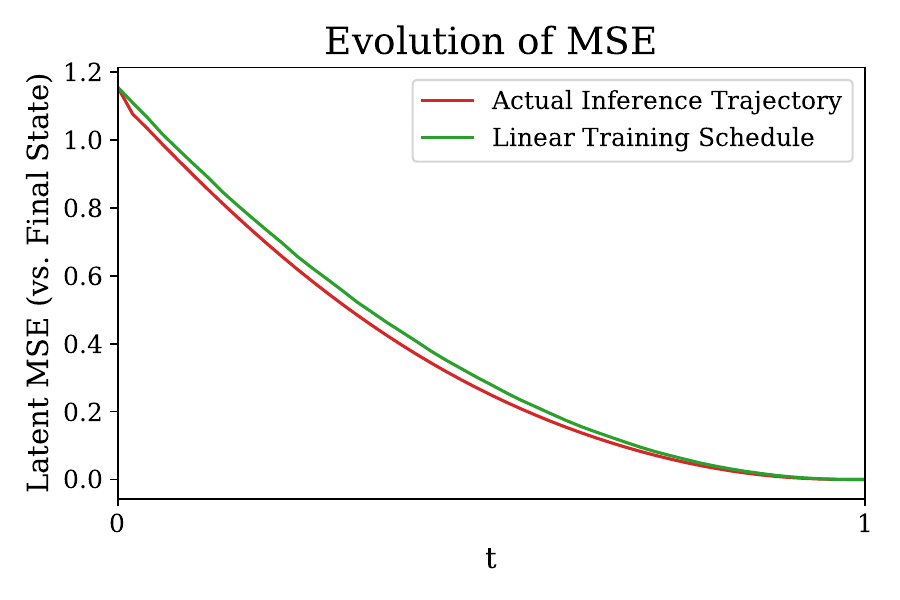} 
        \caption{Latent MSE vs. timestep.}
        \label{fig:analysis_snr_2}
    \end{subfigure}
    \hfill
    % 第四幅图
    \begin{subfigure}[b]{0.45\linewidth}
        \centering
        \includegraphics[width=\linewidth]{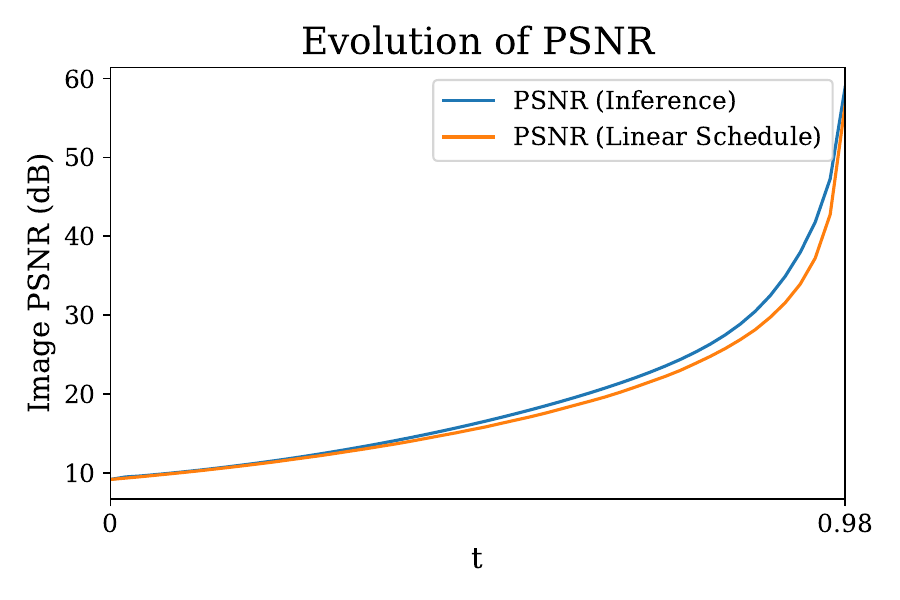}
        \caption{PSNR vs. timestep.}
        \label{fig:analysis_fft_2}
    \end{subfigure}
    
    \caption{Trajectory analysis on GenAI-Bench: Latent MSE (left) and PSNR (right) evolution.}
    \label{fig:analysis_dynamics}
\end{figure}

Our experiments reveal that  time-wise fusion  leads to degraded generation quality, manifesting as blurriness and detail loss. We attribute this failure to a fundamental train--inference instability regarding the evolution of the SNR.

\paragraph{Empirical Evidence on GenAI-Bench.}
To quantify diffusion dynamics, we analyze the divergence between the actual inference trajectory and the theoretical forward process on GenAI-Bench. We generate images using a CFG scale of 6.0 and 50 steps. By constructing a reference trajectory from the final generated latent $\hat{x}_1$ via the training forward process, we evaluate the deviation of the intermediate latent $x_t$. As shown in Figure~\ref{fig:analysis_dynamics}, the actual inference trajectory consistently outpaces the training schedule, exhibiting lower MSE and higher PSNR at identical nominal timesteps. This effectively decouples the real SNR from the nominal timestep $t$.

\paragraph{Mechanism: Iterative vs. Static Sampling.}
The root cause of this drift lies in the distinct data-construction mechanisms. During training, $x_t$ is sampled from a pre-defined static interpolation, ensuring a rigid bijection between $t$ and SNR. Conversely, inference is an iterative process where $x_t$ is the recursive output of preceding steps. Under CFG, which sharpens the vector field, the model restores structural information faster than the linear training assumption predicts, causing the cumulative denoising progress to run ahead of the fixed schedule.

\paragraph{Consequence: Semantic Lag and Joint Stability.}
This misalignment renders the time-wise gating network $g_\psi(t)$ ineffective. Conditioned solely on the nominal $t$, the gate remains oblivious to the fact that the latent $x_t$ has already reached a ``cleaner" state (semantic lag), thus rigidly injecting coarse-grained training priors that hamper high-frequency detail formation. 

In contrast, the joint strategy (S3) mitigates this by exhibiting  higher temporal stability. As visualized in Figure~\ref{fig:w1_comparison}, S3 weights undergo much smaller variations across timesteps than the S1 weights. This inherent stability, achieved by coupling time with depth, effectively dampens the synchronization errors caused by trajectory misalignment, explaining why S3 remains robust while S1 falters.

\paragraph{Counterfactual Validation: Shifted Timestep.}
To rigorously verify this hypothesis, we conduct a counterfactual experiment: we artificially advance the timestep input to the TCFG to    ``catch up" with the accelerated inference trajectory. We introduce a heuristic shift function modulated by a cosine window active in $t \in (0.2, 1]$:
\begin{equation}
    \label{eq:shift}
    t' = t + \delta(t), \quad 
    \delta(t) = 0.01 \cdot \left(1 - \cos\left(\pi \cdot \frac{t - 0.2}{0.8}\right)\right).
\end{equation}
As shown in Table~\ref{tab:ablation_shift}, this simple calibration yields consistent performance recovery across all metrics (e.g., +0.24 on GenEval). Crucially, this positive signal validates the mechanism of our mismatch hypothesis. However, such a rigid manual shift is evidently insufficient to perfectly rectify the complex nonlinear deviations of the inference trajectory. This limitation suggests that while the diagnosis is correct, developing a truly robust time-aware fusion strategy remains an open challenge for future exploration.

\begin{table}[ht]
  \centering
  \caption{Effect of manual timestep recalibration on S1. The slight recovery validates the mismatch mechanism.}
  \label{tab:ablation_shift}
  \setlength{\tabcolsep}{8pt}
  \small
  \begin{tabular}{l c c c}
    \toprule
    Method & {GenEval $\uparrow$} & {GenAI $\uparrow$} & {UnifiedReward $\uparrow$} \\
    \midrule
    S1 & 63.41 & 76.20 & 2.97 \\
    S1 + Shift & \textbf{63.65} & \textbf{76.46} & \textbf{2.98} \\
    \bottomrule
  \end{tabular}
\end{table}

\subsection{Compute Overhead}

% Requires (preamble):
% \usepackage{booktabs}
% \usepackage{siunitx}
% \sisetup{table-number-alignment=center, detect-weight=true, detect-inline-weight=math}

\begin{table}[!htb]
  \centering
  \caption{Overhead of fusion strategies (relative to the DiT backbone). Lower is better. Values are rounded to the nearest integer.}
  \label{tab:overhead}
  \setlength{\tabcolsep}{6pt}
  \small
  \begin{tabular}{l
                  S[table-format=4.0]
                  S[table-format=3.0]
                  S[table-format=4.0]}
    \toprule
    Method & {Params (M)$\downarrow$} & {FLOPs (T)$\downarrow$} & {Latency (ms)$\downarrow$} \\
    \midrule

    \multicolumn{4}{l}{\textit{Baselines}} \\
    B1: Penult.  & 2247 & 454 & \textbf{2339} \\
    B2: Uniform  & 2247 & 454 & 2370 \\
    B3: Static   & 2247 & 454 & 2373 \\
    \midrule

    \multicolumn{4}{l}{\textit{Deep-fusion baseline}} \\
    FuseDiT      & \textbf{1712} & \textbf{357} & 2575 \\
    \midrule

    \multicolumn{4}{l}{\textit{Our fusion strategies}} \\
    S1: Time     & 2247 & 454 & 2391 \\
    S2: Depth    & 2247 & 470 & 2515 \\
    S3: Joint    & 2249 & 470 & 2523 \\
    \bottomrule
  \end{tabular}
\end{table}

%We summarize the inference-time end-to-end latency and FLOPs in Table~\ref{tab:overhead}. Compared to the standard multi-layer baselines (B1--B3), our proposed adaptive strategies (S1--S3) introduce a marginal overhead of \textbf{less than 4\%} in terms of inference latency. This confirms that the gating MLP modules are highly lightweight and do not constitute a significant computational bottleneck in the overall system.

%\paragraph{Comparison with FuseDiT.} 
%FuseDiT achieves superior efficiency by directly reusing internal LLM K/V states to bypass feature aggregation. However, Table ~\ref{tab:main_results} shows this architectural shortcut significantly degrades generative quality. In contrast, our modular routing prioritizes semantic integrity; by maintaining the full LLM hierarchy, it captures high-order semantic signals lost in tightly coupled schemes.

%\paragraph{Conclusion on Efficiency.} 
%While our method incurs higher computational costs than aggressive deep-fusion architectures, it offers a more effective balance between generation quality and model complexity. The performance gains in complex reasoning tasks justify the retention of the full LLM hierarchy, establishing our approach as a robust solution for high-fidelity synthesis where semantic precision is paramount.

% ---------------- Overhead & FuseDiT comparison (revised) ----------------
We summarize model complexity and inference overhead in Table~\ref{tab:overhead}. Relative to standard baselines (B1--B3), our adaptive variants (S1--S3) add negligible cost. In particular, the best depth-wise strategy (S2) introduces essentially no extra parameters, increases end-to-end latency by only $\sim$8\%, indicating that gating is not a computational bottleneck.

FuseDiT attains lower FLOPs primarily by reusing LLM hidden states and adopting a lighter self-attention design. As shown in Table~\ref{tab:main_results}, this reduction in FLOPs is accompanied by a clear degradation in generative quality. We hypothesize that reusing LLM K/V states,  may restrict conditioning expressiveness  and limit the model's ability to adapt semantic cues. In contrast, our modular routing preserves semantic expressiveness and yields a more favorable quality efficiency trade-off under comparable latency.
\section{Conclusion}

This paper investigates how to  leverage multi-layer LLM representations for text conditioning in DiT-based generative models. We propose a unified fusion formulation that supports controlled implementation and fair comparison of time-wise adaptive fusion, depth-wise adaptive fusion, and their combinations within a single framework. Experiments show that multi-layer fusion consistently outperforms  single-layer conditioning, and that depth-wise adaptive fusion delivers the most robust and substantial improvements among the studied strategies. By contrast, purely time-wise fusion can hurt performance, which we attribute to a train–inference mismatch between nominal timesteps and inference-time denoising dynamics. Interestingly, the learned time-wise weights remain structured across timesteps, suggesting that effective time-adaptive conditioning may be possible when driven by trajectory-aligned signals.    
\section{Impact Statements}

This paper aims to advance research in machine learning and generative modeling. We introduce a semantic routing mechanism for diffusion Transformers that performs lightweight and interpretable fusion of multi layer large language model hidden states, better matching the generation process across network depth and optional temporal dynamics, thereby improving text image alignment and compositional instruction following.

Potential positive impacts include improved semantic alignment between generated content and textual inputs, enabled by stronger textual conditioning. This can enhance controllability and instruction consistency in text to image generation, benefiting applications such as controllable synthesis, content editing, and human AI co creation. The proposed lightweight gated fusion also provides a more interpretable interface for analysis, which can support scientific understanding of how textual conditioning operates during generation and motivate more robust conditioning designs.

As with many advances in high fidelity text to image generation, our approach could be misused to produce misleading or deceptive imagery, potentially amplifying misinformation. Moreover, if training data or the underlying text encoder contains societal biases, stronger alignment may reproduce such biases more consistently and could exacerbate stereotyping or unfair representations. Enhanced compliance with fine grained prompts may also lower the barrier to generating harmful or infringing content.

We do not expect this work to introduce fundamentally new categories of safety risks. In deployment or release settings, existing safety and compliance practices commonly used for generative models can still be applied to mitigate potential misuse, bias, and infringement risks, such as content safety filtering, sensitive concept blocking, bias and robustness evaluation, and, when appropriate, provenance mechanisms including watermarking.

%\bibliography{main}

\begin{thebibliography}{47}
\providecommand{\natexlab}[1]{#1}
\providecommand{\url}[1]{\texttt{#1}}
\expandafter\ifx\csname urlstyle\endcsname\relax
  \providecommand{\doi}[1]{doi: #1}\else
  \providecommand{\doi}{doi: \begingroup \urlstyle{rm}\Url}\fi

\bibitem[Ba et~al.(2016)Ba, Kiros, and Hinton]{ba2016layer}
Ba, J.~L., Kiros, J.~R., and Hinton, G.~E.
\newblock Layer normalization.
\newblock \emph{arXiv preprint arXiv:1607.06450}, 2016.

\bibitem[Bai et~al.(2025)Bai, Cai, Chen, Chen, Chen, Cheng, Deng, Ding, Gao, Ge, Ge, Guo, Huang, Huang, Huang, Hui, Jiang, Li, Li, Li, Li, Lin, Lin, Liu, Liu, Liu, Liu, Liu, Liu, Lu, Luo, Lv, Men, Meng, Ren, Ren, Song, Sun, Tang, Tu, Wan, Wang, Wang, Wang, Wang, Xie, Xu, Xu, Xu, Yang, Yang, Yang, Yang, Yu, Zhang, Zhang, Zhang, Zheng, Zhong, Zhou, Zhou, Zhou, Zhu, and Zhu]{Qwen3-VL}
Bai, S., Cai, Y., Chen, R., Chen, K., Chen, X., Cheng, Z., Deng, L., Ding, W., Gao, C., Ge, C., Ge, W., Guo, Z., Huang, Q., Huang, J., Huang, F., Hui, B., Jiang, S., Li, Z., Li, M., Li, M., Li, K., Lin, Z., Lin, J., Liu, X., Liu, J., Liu, C., Liu, Y., Liu, D., Liu, S., Lu, D., Luo, R., Lv, C., Men, R., Meng, L., Ren, X., Ren, X., Song, S., Sun, Y., Tang, J., Tu, J., Wan, J., Wang, P., Wang, P., Wang, Q., Wang, Y., Xie, T., Xu, Y., Xu, H., Xu, J., Yang, Z., Yang, M., Yang, J., Yang, A., Yu, B., Zhang, F., Zhang, H., Zhang, X., Zheng, B., Zhong, H., Zhou, J., Zhou, F., Zhou, J., Zhu, Y., and Zhu, K.
\newblock Qwen3-vl technical report.
\newblock \emph{arXiv preprint arXiv:2511.21631}, 2025.

\bibitem[Barbero et~al.(2025)Barbero, Arroyo, Gu, Perivolaropoulos, Bronstein, Veli{\v{c}}kovi{\'c}, and Pascanu]{barbero2025llms}
Barbero, F., Arroyo, A., Gu, X., Perivolaropoulos, C., Bronstein, M., Veli{\v{c}}kovi{\'c}, P., and Pascanu, R.
\newblock Why do llms attend to the first token?
\newblock \emph{arXiv preprint arXiv:2504.02732}, 2025.

\bibitem[BehnamGhader et~al.(2024)BehnamGhader, Adlakha, Mosbach, Bahdanau, Chapados, and Reddy]{behnamghader2024llm2vec}
BehnamGhader, P., Adlakha, V., Mosbach, M., Bahdanau, D., Chapados, N., and Reddy, S.
\newblock Llm2vec: Large language models are secretly powerful text encoders.
\newblock \emph{arXiv preprint arXiv:2404.05961}, 2024.

\bibitem[Cai et~al.(2025)Cai, Cao, Du, Gao, Hoi, Huang, Hou, Jiang, Jin, Li, et~al.]{cai2025z}
Cai, H., Cao, S., Du, R., Gao, P., Hoi, S., Huang, S., Hou, Z., Jiang, D., Jin, X., Li, L., et~al.
\newblock Z-image: An efficient image generation foundation model with single-stream diffusion transformer.
\newblock \emph{arXiv preprint arXiv:2511.22699}, 2025.

\bibitem[Chen et~al.(2023)Chen, Yu, Ge, Yao, Xie, Wu, Wang, Kwok, Luo, Lu, and Li]{chen2023pixart}
Chen, J., Yu, J., Ge, C., Yao, L., Xie, E., Wu, Y., Wang, Z., Kwok, J., Luo, P., Lu, H., and Li, Z.
\newblock Pixart-$\alpha$: Fast training of diffusion transformer for photorealistic text-to-image synthesis, 2023.
\newblock URL \url{https://arxiv.org/abs/2310.00426}.

\bibitem[Chen et~al.(2024)Chen, Shen, Ye, Cao, Tu, Bouganis, Zhao, and Chen]{chen2024deltadittrainingfreeaccelerationmethod}
Chen, P., Shen, M., Ye, P., Cao, J., Tu, C., Bouganis, C.-S., Zhao, Y., and Chen, T.
\newblock $\delta$-dit: A training-free acceleration method tailored for diffusion transformers.
\newblock \emph{ArXiv}, abs/2406.01125, 2024.
\newblock URL \url{https://api.semanticscholar.org/CorpusID:270215326}.

\bibitem[Esser et~al.(2024)Esser, Kulal, Blattmann, Entezari, M{\"u}ller, Saini, Levi, Lorenz, Sauer, Boesel, et~al.]{esser2024scaling}
Esser, P., Kulal, S., Blattmann, A., Entezari, R., M{\"u}ller, J., Saini, H., Levi, Y., Lorenz, D., Sauer, A., Boesel, F., et~al.
\newblock Scaling rectified flow transformers for high-resolution image synthesis.
\newblock In \emph{Forty-first international conference on machine learning}, 2024.

\bibitem[Fan et~al.(2024)Fan, Jiang, Li, Meng, Han, Shang, Sun, Wang, and Wang]{fan2024not}
Fan, S., Jiang, X., Li, X., Meng, X., Han, P., Shang, S., Sun, A., Wang, Y., and Wang, Z.
\newblock Not all layers of llms are necessary during inference.
\newblock \emph{arXiv preprint arXiv:2403.02181}, 2024.

\bibitem[Gao et~al.(2025)Gao, Guo, Hoang, Huang, Jiang, Kong, Li, Li, Li, Li, et~al.]{gao2025seedance}
Gao, Y., Guo, H., Hoang, T., Huang, W., Jiang, L., Kong, F., Li, H., Li, J., Li, L., Li, X., et~al.
\newblock Seedance 1.0: Exploring the boundaries of video generation models.
\newblock \emph{arXiv preprint arXiv:2506.09113}, 2025.

\bibitem[Ghosh et~al.(2023)Ghosh, Hajishirzi, and Schmidt]{ghosh2023geneval}
Ghosh, D., Hajishirzi, H., and Schmidt, L.
\newblock Geneval: An object-focused framework for evaluating text-to-image alignment.
\newblock \emph{Advances in Neural Information Processing Systems}, 36:\penalty0 52132--52152, 2023.

\bibitem[Gong et~al.(2022)Gong, Chen, Chen, and Wang]{gong2022sandwich}
Gong, X., Chen, W., Chen, T., and Wang, Z.
\newblock Sandwich batch normalization: A drop-in replacement for feature distribution heterogeneity.
\newblock In \emph{Proceedings of the IEEE/CVF Winter Conference on Applications of Computer Vision}, pp.\  2494--2504, 2022.

\bibitem[Gurnee \& Tegmark(2023)Gurnee and Tegmark]{gurnee2023language}
Gurnee, W. and Tegmark, M.
\newblock Language models represent space and time.
\newblock \emph{arXiv preprint arXiv:2310.02207}, 2023.

\bibitem[Henry et~al.(2020)Henry, Dachapally, Pawar, and Chen]{henry2020query}
Henry, A., Dachapally, P.~R., Pawar, S.~S., and Chen, Y.
\newblock Query-key normalization for transformers.
\newblock In \emph{Findings of the Association for Computational Linguistics: EMNLP 2020}, pp.\  4246--4253, 2020.

\bibitem[Heo et~al.(2024)Heo, Park, Han, and Yun]{heo2024rotary}
Heo, B., Park, S., Han, D., and Yun, S.
\newblock Rotary position embedding for vision transformer.
\newblock In \emph{European Conference on Computer Vision}, pp.\  289--305. Springer, 2024.

\bibitem[Hertz et~al.(2022)Hertz, Mokady, Tenenbaum, Aberman, Pritch, and Cohen-Or]{hertz2022prompt}
Hertz, A., Mokady, R., Tenenbaum, J., Aberman, K., Pritch, Y., and Cohen-Or, D.
\newblock Prompt-to-prompt image editing with cross attention control.
\newblock \emph{arXiv preprint arXiv:2208.01626}, 2022.

\bibitem[Ho \& Salimans(2022)Ho and Salimans]{ho2022classifier}
Ho, J. and Salimans, T.
\newblock Classifier-free diffusion guidance.
\newblock \emph{arXiv preprint arXiv:2207.12598}, 2022.

\bibitem[Jin et~al.(2025)Jin, Yu, Huang, Zeng, Wang, Hua, Zhao, Mei, Meng, Ding, et~al.]{jin2025exploring}
Jin, M., Yu, Q., Huang, J., Zeng, Q., Wang, Z., Hua, W., Zhao, H., Mei, K., Meng, Y., Ding, K., et~al.
\newblock Exploring concept depth: How large language models acquire knowledge and concept at different layers?
\newblock In \emph{Proceedings of the 31st International Conference on Computational Linguistics}, pp.\  558--573, 2025.

\bibitem[Kim et~al.(2025)Kim, Lee, Park, Oh, Kim, Yoo, Shin, Han, Shin, and Yoo]{kim2025peri}
Kim, J., Lee, B., Park, C., Oh, Y., Kim, B., Yoo, T., Shin, S., Han, D., Shin, J., and Yoo, K.~M.
\newblock Peri-ln: Revisiting normalization layer in the transformer architecture.
\newblock \emph{arXiv preprint arXiv:2502.02732}, 2025.

\bibitem[Kong et~al.(2024)Kong, Tian, Zhang, Min, Dai, Zhou, Xiong, Li, Wu, Zhang, et~al.]{kong2024hunyuanvideo}
Kong, W., Tian, Q., Zhang, Z., Min, R., Dai, Z., Zhou, J., Xiong, J., Li, X., Wu, B., Zhang, J., et~al.
\newblock Hunyuanvideo: A systematic framework for large video generative models.
\newblock \emph{arXiv preprint arXiv:2412.03603}, 2024.

\bibitem[Li et~al.(2024)Li, Lin, Pathak, Li, Fei, Wu, Ling, Xia, Zhang, Neubig, et~al.]{li2024genai}
Li, B., Lin, Z., Pathak, D., Li, J., Fei, Y., Wu, K., Ling, T., Xia, X., Zhang, P., Neubig, G., et~al.
\newblock Genai-bench: Evaluating and improving compositional text-to-visual generation.
\newblock \emph{arXiv preprint arXiv:2406.13743}, 2024.

\bibitem[Li et~al.(2025{\natexlab{a}})Li, Xue, Yang, Shi, Chen, Guan, Zhang, and Zhang]{li2025unseen}
Li, B., Xue, X., Yang, S., Shi, Y., Chen, X., Guan, Y., Zhang, Y., and Zhang, W.
\newblock The unseen bias: How norm discrepancy in pre-norm mllms leads to visual information loss.
\newblock \emph{arXiv preprint arXiv:2512.08374}, 2025{\natexlab{a}}.

\bibitem[Li et~al.(2025{\natexlab{b}})Li, Yang, Guan, An, Chen, Shi, Wan, Zhang, et~al.]{li2025gran}
Li, B., Yang, S., Guan, Y., An, R., Chen, X., Shi, Y., Wan, P., Zhang, W., et~al.
\newblock Gran-ted: Generating robust, aligned, and nuanced text embedding for diffusion models.
\newblock \emph{arXiv preprint arXiv:2512.15560}, 2025{\natexlab{b}}.

\bibitem[Lipman et~al.(2022)Lipman, Chen, Ben-Hamu, Nickel, and Le]{lipman2022flow}
Lipman, Y., Chen, R.~T., Ben-Hamu, H., Nickel, M., and Le, M.
\newblock Flow matching for generative modeling.
\newblock \emph{arXiv preprint arXiv:2210.02747}, 2022.

\bibitem[Liu et~al.(2024{\natexlab{a}})Liu, Akhgari, Visheratin, Kamko, Xu, Shrirao, Lambert, Souza, Doshi, and Li]{liu2024playground}
Liu, B., Akhgari, E., Visheratin, A., Kamko, A., Xu, L., Shrirao, S., Lambert, C., Souza, J., Doshi, S., and Li, D.
\newblock Playground v3: Improving text-to-image alignment with deep-fusion large language models.
\newblock \emph{arXiv preprint arXiv:2409.10695}, 2024{\natexlab{a}}.

\bibitem[Liu et~al.(2023)Liu, Ning, Lin, Yang, and Wang]{liu2023oms}
Liu, E., Ning, X., Lin, Z., Yang, H., and Wang, Y.
\newblock Oms-dpm: Optimizing the model schedule for diffusion probabilistic models.
\newblock In \emph{International Conference on Machine Learning}, pp.\  21915--21936. PMLR, 2023.

\bibitem[Liu et~al.(2024{\natexlab{b}})Liu, Kong, Liu, and Sun]{liu2024fantastic}
Liu, Z., Kong, C., Liu, Y., and Sun, M.
\newblock Fantastic semantics and where to find them: Investigating which layers of generative llms reflect lexical semantics.
\newblock \emph{arXiv preprint arXiv:2403.01509}, 2024{\natexlab{b}}.

\bibitem[Loshchilov \& Hutter(2017)Loshchilov and Hutter]{loshchilov2017decoupled}
Loshchilov, I. and Hutter, F.
\newblock Decoupled weight decay regularization.
\newblock \emph{arXiv preprint arXiv:1711.05101}, 2017.

\bibitem[Ma et~al.(2024)Ma, Zong, Song, Li, and Liu]{ma2024exploring}
Ma, B., Zong, Z., Song, G., Li, H., and Liu, Y.
\newblock Exploring the role of large language models in prompt encoding for diffusion models.
\newblock \emph{arXiv preprint arXiv:2406.11831}, 2024.

\bibitem[Ma et~al.(2025)Ma, Huang, Yan, Chen, Duan, Yin, Wan, Ming, Song, Chen, et~al.]{ma2025step}
Ma, G., Huang, H., Yan, K., Chen, L., Duan, N., Yin, S., Wan, C., Ming, R., Song, X., Chen, X., et~al.
\newblock Step-video-t2v technical report: The practice, challenges, and future of video foundation model.
\newblock \emph{arXiv preprint arXiv:2502.10248}, 2025.

\bibitem[Peebles \& Xie(2023)Peebles and Xie]{peebles2023scalable}
Peebles, W. and Xie, S.
\newblock Scalable diffusion models with transformers.
\newblock In \emph{Proceedings of the IEEE/CVF international conference on computer vision}, pp.\  4195--4205, 2023.

\bibitem[Radford et~al.(2021)Radford, Kim, Hallacy, Ramesh, Goh, Agarwal, Sastry, Askell, Mishkin, Clark, et~al.]{radford2021learning}
Radford, A., Kim, J.~W., Hallacy, C., Ramesh, A., Goh, G., Agarwal, S., Sastry, G., Askell, A., Mishkin, P., Clark, J., et~al.
\newblock Learning transferable visual models from natural language supervision.
\newblock In \emph{International conference on machine learning}, pp.\  8748--8763. PmLR, 2021.

\bibitem[Raffel et~al.(2020)Raffel, Shazeer, Roberts, Lee, Narang, Matena, Zhou, Li, and Liu]{raffel2020exploring}
Raffel, C., Shazeer, N., Roberts, A., Lee, K., Narang, S., Matena, M., Zhou, Y., Li, W., and Liu, P.~J.
\newblock Exploring the limits of transfer learning with a unified text-to-text transformer.
\newblock \emph{Journal of machine learning research}, 21\penalty0 (140):\penalty0 1--67, 2020.

\bibitem[Ronneberger et~al.(2015)Ronneberger, Fischer, and Brox]{ronneberger2015u}
Ronneberger, O., Fischer, P., and Brox, T.
\newblock U-net: Convolutional networks for biomedical image segmentation.
\newblock In \emph{International Conference on Medical image computing and computer-assisted intervention}, pp.\  234--241. Springer, 2015.

\bibitem[Saharia et~al.(2022)Saharia, Chan, Saxena, Li, Whang, Denton, Ghasemipour, Gontijo~Lopes, Karagol~Ayan, Salimans, et~al.]{saharia2022photorealistic}
Saharia, C., Chan, W., Saxena, S., Li, L., Whang, J., Denton, E.~L., Ghasemipour, K., Gontijo~Lopes, R., Karagol~Ayan, B., Salimans, T., et~al.
\newblock Photorealistic text-to-image diffusion models with deep language understanding.
\newblock \emph{Advances in neural information processing systems}, 35:\penalty0 36479--36494, 2022.

\bibitem[Schuhmann et~al.(2021)Schuhmann, Vencu, Beaumont, Kaczmarczyk, Mullis, Katta, Coombes, Jitsev, and Komatsuzaki]{schuhmann2021laion}
Schuhmann, C., Vencu, R., Beaumont, R., Kaczmarczyk, R., Mullis, C., Katta, A., Coombes, T., Jitsev, J., and Komatsuzaki, A.
\newblock Laion-400m: Open dataset of clip-filtered 400 million image-text pairs.
\newblock \emph{arXiv preprint arXiv:2111.02114}, 2021.

\bibitem[Seedream et~al.(2025)Seedream, Chen, Gao, Gong, Guo, Guo, Guo, Hou, Huang, Huang, et~al.]{seedream2025seedream}
Seedream, T., Chen, Y., Gao, Y., Gong, L., Guo, M., Guo, Q., Guo, Z., Hou, X., Huang, W., Huang, Y., et~al.
\newblock Seedream 4.0: Toward next-generation multimodal image generation.
\newblock \emph{arXiv preprint arXiv:2509.20427}, 2025.

\bibitem[Skean et~al.(2025)Skean, Arefin, Zhao, Patel, Naghiyev, LeCun, and Shwartz-Ziv]{skean2025layer}
Skean, O., Arefin, M.~R., Zhao, D., Patel, N., Naghiyev, J., LeCun, Y., and Shwartz-Ziv, R.
\newblock Layer by layer: Uncovering hidden representations in language models.
\newblock \emph{arXiv preprint arXiv:2502.02013}, 2025.

\bibitem[Su et~al.(2024)Su, Ahmed, Lu, Pan, Bo, and Liu]{su2024roformer}
Su, J., Ahmed, M., Lu, Y., Pan, S., Bo, W., and Liu, Y.
\newblock Roformer: Enhanced transformer with rotary position embedding.
\newblock \emph{Neurocomputing}, 568:\penalty0 127063, 2024.

\bibitem[Tang et~al.(2025)Tang, Zheng, Paul, and Xie]{tang2025exploring}
Tang, B., Zheng, B., Paul, S., and Xie, S.
\newblock Exploring the deep fusion of large language models and diffusion transformers for text-to-image synthesis.
\newblock In \emph{Proceedings of the Computer Vision and Pattern Recognition Conference}, pp.\  28586--28595, 2025.

\bibitem[Touvron et~al.(2023)Touvron, Martin, Stone, Albert, Almahairi, Babaei, Bashlykov, Batra, Bhargava, Bhosale, et~al.]{touvron2023llama}
Touvron, H., Martin, L., Stone, K., Albert, P., Almahairi, A., Babaei, Y., Bashlykov, N., Batra, S., Bhargava, P., Bhosale, S., et~al.
\newblock Llama 2: Open foundation and fine-tuned chat models.
\newblock \emph{arXiv preprint arXiv:2307.09288}, 2023.

\bibitem[von Platen et~al.(2022)von Platen, Patil, Lozhkov, Cuenca, Lambert, Rasul, Davaadorj, Nair, Paul, Berman, Xu, Liu, and Wolf]{von-platen-etal-2022-diffusers}
von Platen, P., Patil, S., Lozhkov, A., Cuenca, P., Lambert, N., Rasul, K., Davaadorj, M., Nair, D., Paul, S., Berman, W., Xu, Y., Liu, S., and Wolf, T.
\newblock Diffusers: State-of-the-art diffusion models.
\newblock \url{https://github.com/huggingface/diffusers}, 2022.

\bibitem[Wan et~al.(2025)Wan, Wang, Ai, Wen, Mao, Xie, Chen, Yu, Zhao, Yang, et~al.]{wan2025wan}
Wan, T., Wang, A., Ai, B., Wen, B., Mao, C., Xie, C.-W., Chen, D., Yu, F., Zhao, H., Yang, J., et~al.
\newblock Wan: Open and advanced large-scale video generative models.
\newblock \emph{arXiv preprint arXiv:2503.20314}, 2025.

\bibitem[Wang et~al.(2025{\natexlab{a}})Wang, Ge, Karras, Liu, and Balaji]{wang2025comprehensive}
Wang, A.~Z., Ge, S., Karras, T., Liu, M.-Y., and Balaji, Y.
\newblock A comprehensive study of decoder-only llms for text-to-image generation.
\newblock In \emph{Proceedings of the Computer Vision and Pattern Recognition Conference}, pp.\  28575--28585, 2025{\natexlab{a}}.

\bibitem[Wang \& Vastola(2023)Wang and Vastola]{wang2023diffusion}
Wang, B. and Vastola, J.~J.
\newblock Diffusion models generate images like painters: an analytical theory of outline first, details later.
\newblock \emph{arXiv preprint arXiv:2303.02490}, 2023.

\bibitem[Wang et~al.(2025{\natexlab{b}})Wang, Zang, Li, Jin, and Wang]{unifiedreward}
Wang, Y., Zang, Y., Li, H., Jin, C., and Wang, J.
\newblock Unified reward model for multimodal understanding and generation.
\newblock \emph{arXiv preprint arXiv:2503.05236}, 2025{\natexlab{b}}.

\bibitem[Wu et~al.(2025)Wu, Li, Zhou, Lin, Gao, Yan, Yin, Bai, Xu, Chen, et~al.]{wu2025qwen}
Wu, C., Li, J., Zhou, J., Lin, J., Gao, K., Yan, K., Yin, S.-m., Bai, S., Xu, X., Chen, Y., et~al.
\newblock Qwen-image technical report.
\newblock \emph{arXiv preprint arXiv:2508.02324}, 2025.

\end{thebibliography}
%\bibliographystyle{icml2026}

%%%%%%%%%%%%%%%%%%%%%%%%%%%%%%%%%%%%%%%%%%%%%%%%%%%%%%%%%%%%%%%%%%%%%%%%%%%%%%%
%%%%%%%%%%%%%%%%%%%%%%%%%%%%%%%%%%%%%%%%%%%%%%%%%%%%%%%%%%%%%%%%%%%%%%%%%%%%%%%
% APPENDIX
%%%%%%%%%%%%%%%%%%%%%%%%%%%%%%%%%%%%%%%%%%%%%%%%%%%%%%%%%%%%%%%%%%%%%%%%%%%%%%%
%%%%%%%%%%%%%%%%%%%%%%%%%%%%%%%%%%%%%%%%%%%%%%%%%%%%%%%%%%%%%%%%%%%%%%%%%%%%%%%

\appendix
\onecolumn

\section{TCFG Implementation Details}
\label{app:TCFG_details}
The Time-Conditioned Fusion Gate (TCFG) serves as the core module for adaptively aggregating multi-layer LLM features based on the diffusion timestep. Our implementation follows a lightweight Multi-Layer Perceptron (MLP) design with specific initialization strategies to ensure training stability.

\paragraph{Sinusoidal Timestep Embedding.}
The continuous timestep $t$ is first mapped to a high-dimensional feature vector using a sinusoidal embedding, similar to the positional encoding in Transformers. For a time embedding dimension $D_t$, the embedding $\phi(t) \in \mathbb{R}^{D_t}$ is computed as:
\begin{equation}
    \phi(t) = \left[ \dots, \cos(t \cdot \omega_i), \sin(t \cdot \omega_i), \dots \right], \quad \text{where } \omega_i = \frac{1}{10000^{2i/D_t}}.
\end{equation}
In our experiments, we set $D_t = 128$.

\paragraph{Gating Network Architecture.}
The embedding $\phi(t)$ is processed by a two-layer MLP to generate the fusion logits $z_t \in \mathbb{R}^L$, where $L$ is the number of LLM layers. The network structure consists of:
\begin{enumerate}
    \item \textbf{Input Projection:} A linear layer mapping from $D_t$ to a hidden dimension of $4 \times D_t$.
    \item \textbf{Activation:} A Sigmoid Linear Unit (SiLU) activation function.
    \item \textbf{Output Projection:} A linear layer mapping from $4 \times D_t$ to $L$.
\end{enumerate}

\paragraph{Zero-Initialization Strategy.}
To facilitate a smooth starting point for optimization, we employ a zero-initialization strategy for the final output projection layer. Specifically, both the weight matrix and the bias vector of the second linear layer are initialized to zero. 
Consequently, at the beginning of training, the output logits $z_t$ are zero vectors, which results in a uniform probability distribution after the Softmax operation (i.e., $\alpha_t^{(l)} = 1/L$ for all $l$). This ensures that the model initially utilizes an average of all layers before learning specific routing preferences.

\paragraph{Feature Normalization and Aggregation.}
Given a set of hidden states $\{H^{(l)}\}_{l=1}^L$ from the text encoder, we strictly apply Layer Normalization (LayerNorm) \textit{before} fusion to handle scale discrepancies across layers. The final aggregated feature $H_{\text{cond}}$ is computed as:
\begin{equation}
    H_{\text{cond}} = \sum_{l=1}^L \text{Softmax}(z_t)^{(l)} \cdot \text{LayerNorm}(H^{(l)}).
\end{equation}
This formulation ensures that the aggregated feature remains within the convex hull of the normalized representations, maintaining numerical stability throughout the training process.

\section{Image Examples}
\label{app:img}

\newcommand{\StrategyRowSeven}[7]{%
  \setlength{\tabcolsep}{1.5pt}%
  \renewcommand{\arraystretch}{1.0}%
  \begin{tabular}{ccccccc}
    {\scriptsize Penult} &
    {\scriptsize Uniform} &
    {\scriptsize Static} &
    {\scriptsize FuseDiT} &
    {\scriptsize Time} &
    {\scriptsize Depth} &
    {\scriptsize Joint} \\
    \includegraphics[width=0.135\linewidth]{#1} &
    \includegraphics[width=0.135\linewidth]{#2} &
    \includegraphics[width=0.135\linewidth]{#3} &
    \includegraphics[width=0.135\linewidth]{#4} &
    \includegraphics[width=0.135\linewidth]{#5} &
    \includegraphics[width=0.135\linewidth]{#6} &
    \includegraphics[width=0.135\linewidth]{#7} \\
  \end{tabular}%
}

% 一个prompt面板：#1 prompt文本，#2..#8 为7张图路径
\newcommand{\PromptPanelSeven}[8]{%
  \Needspace{0.26\textheight}% 可调/可删：避免一个面板被断页拆开
  {\small Prompt: ``#1''}\par%
  \StrategyRowSeven{#2}{#3}{#4}{#5}{#6}{#7}{#8}\par%
}

\begin{center}
\captionsetup{type=figure}
\captionof{figure}{Qualitative comparisons across strategies under multiple prompts. Columns: B1/B2/B3, FuseDiT baseline, and three fusion strategies (S1/S2/S3). All images use identical sampling settings for fair comparison.}
\label{fig:appxA_qual_all}

\PromptPanelSeven
  {Among a group of pastel-colored balloons, one stands out in vibrant red.}
  {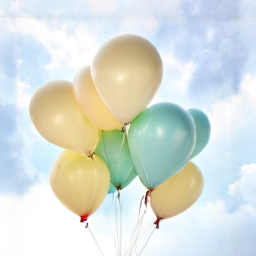}
  {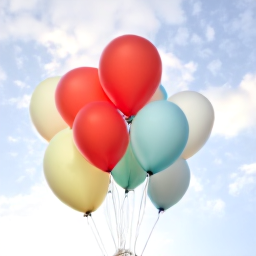}
  {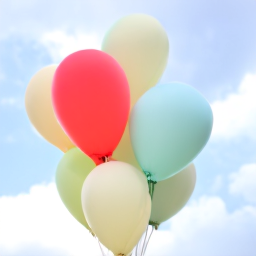}
  {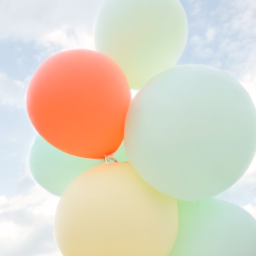}
  {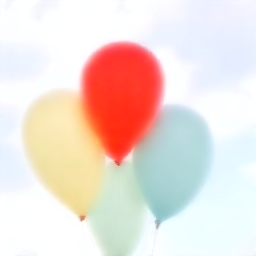}
  {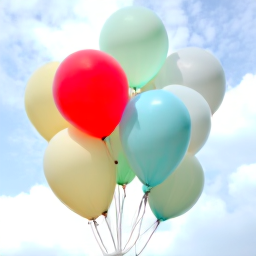}
  {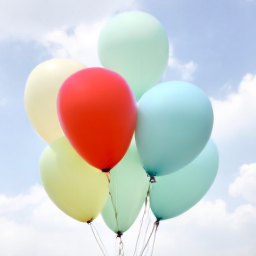}

\PromptPanelSeven
  {A vase with five purple roses on a kitchen table.}
  {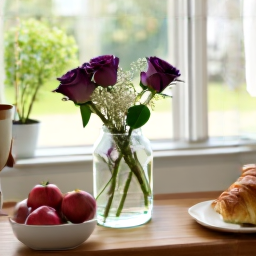} 
  {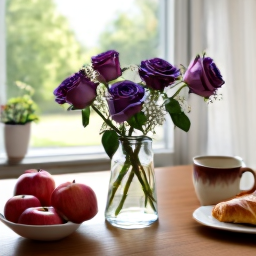}    
  {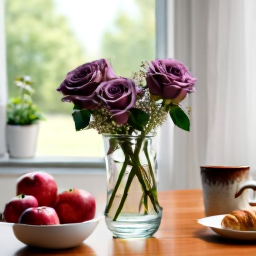} 
  {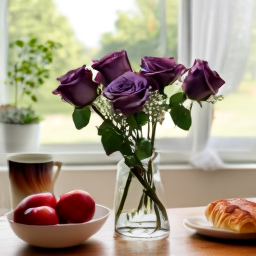}  
  {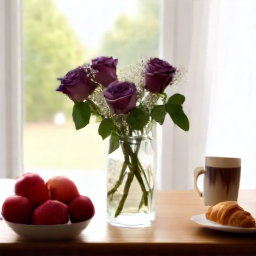}      
  {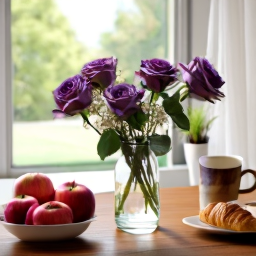}
  {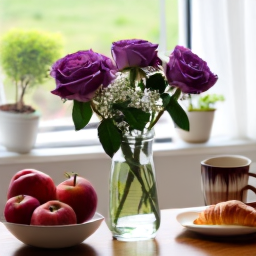}  
  
\PromptPanelSeven
  {A pilot with aviator sunglasses.}
  {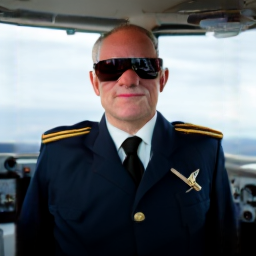}
  {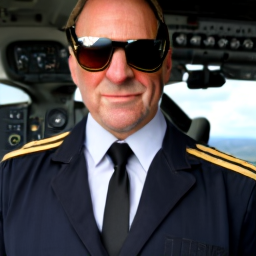}
  {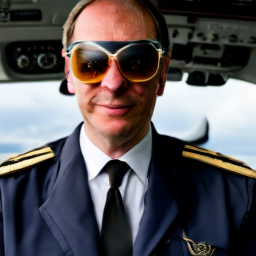}
  {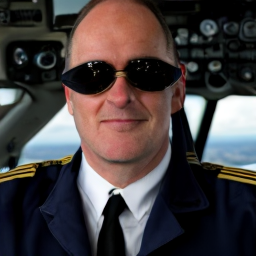}
  {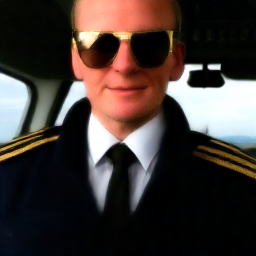}
  {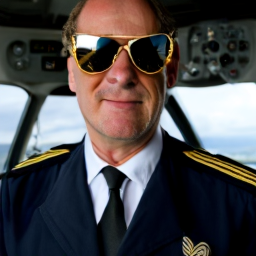}
  {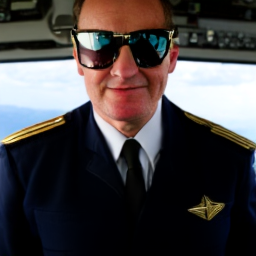}

\PromptPanelSeven
  {Five colorful balloons floating against a clear blue sky.}
  {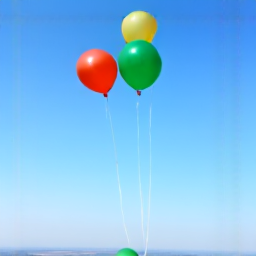} 
  {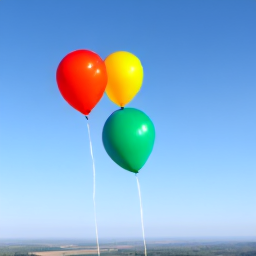}  
  {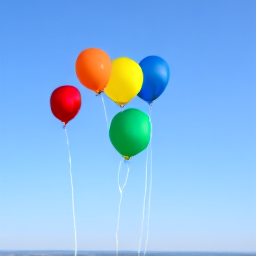}       
  {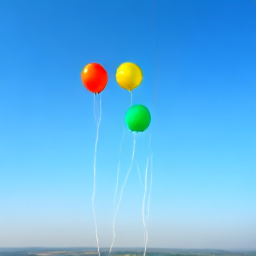}
  {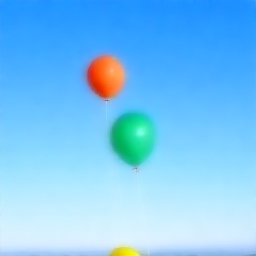}      
  {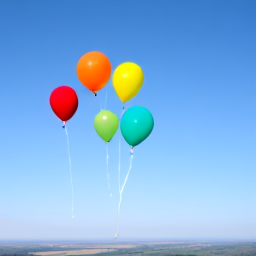}
  {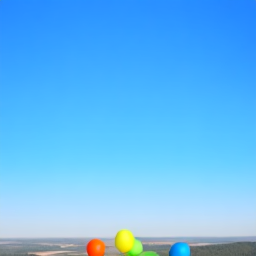}

%\PromptPanelSeven
%  {A colorful skirt has an uncolorful hem.}
%  {example_image/baseline/00192_sample_0.png}
%  {example_image/uniform/00192_sample_0.png}
%  {example_image/static/00192_sample_0.png}   
%  {example_image/fusedit/00192_sample_0.png}
%  {example_image/time-wise/00192_sample_0.png}
%  {example_image/depth-wise/00192_sample_0.png}
%  {example_image/joint/00192_sample_0.png}  

%\PromptPanelSeven
%  {A big green apple next to a small red apple on a kitchen counter.}
%  {example_image/baseline/00385_sample_0.png}
%  {example_image/uniform/00385_sample_0.png}
%  {example_image/static/00385_sample_0.png}
%  {example_image/fusedit/00385_sample_0.png}
%  {example_image/time-wise/00385_sample_0.png}      
%  {example_image/depth-wise/00385_sample_0.png}
%  {example_image/joint/00385_sample_0.png}

\PromptPanelSeven
  {A large pizza with pepperoni on the left half and mushrooms on the right half.}
  {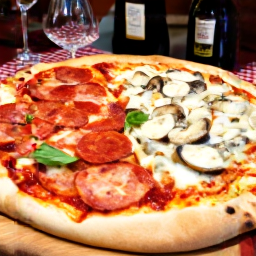}
  {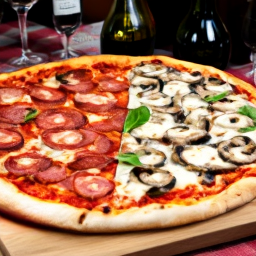}    
  {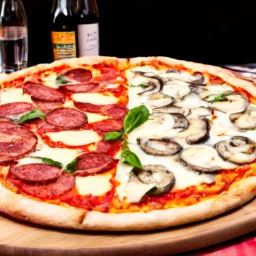}
  {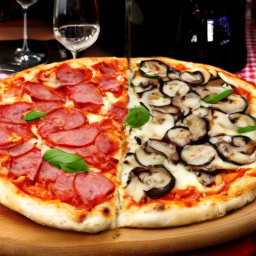}
  {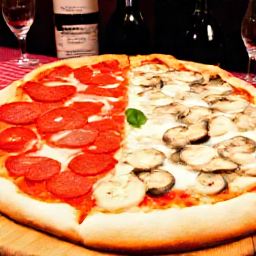}      
  {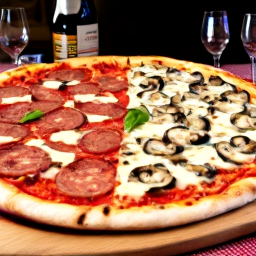}
  {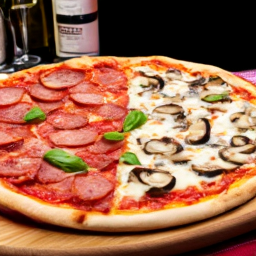}

\PromptPanelSeven
  {A large teddy bear wearing a bow tie next to a small teddy bear wearing a party hat.}
  {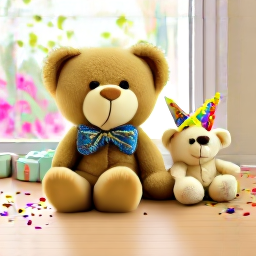}
  {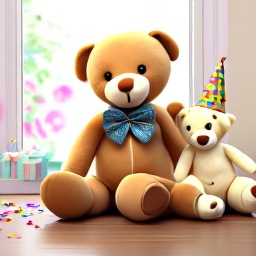}    
  {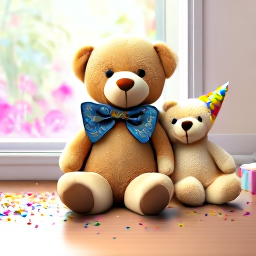}
  {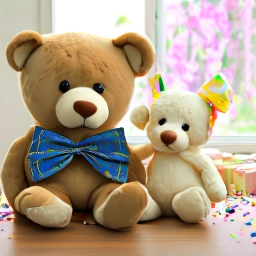}
  {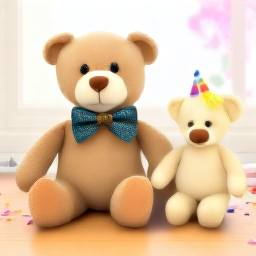}      
  {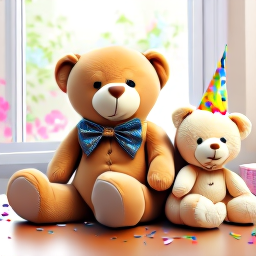}
  {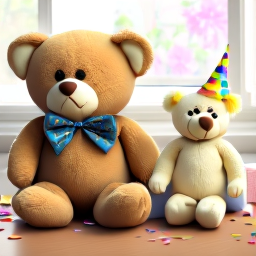}

\PromptPanelSeven
  {A small dog in a cozy orange sweater sitting beside a cat wearing a stylish blue bow tie.}
  {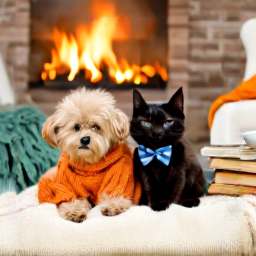}
  {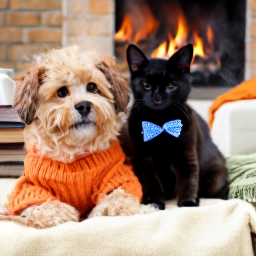}    
  {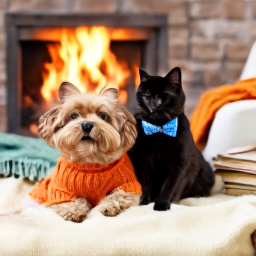} 
  {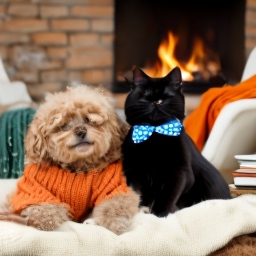}
  {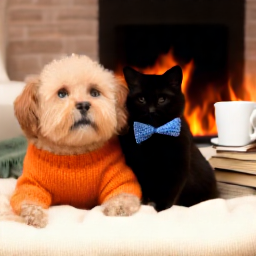}      
  {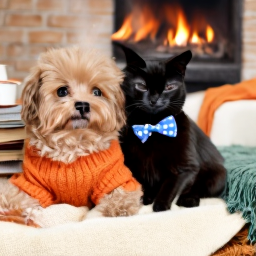}
  {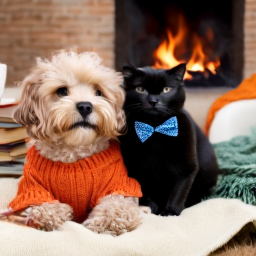}  

\PromptPanelSeven
  {An orange cat lies on a couch surrounded by three pillows that are all blue.}
  {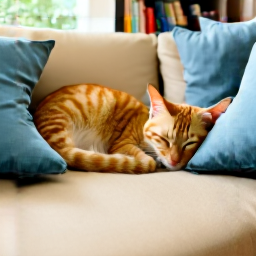} 
  {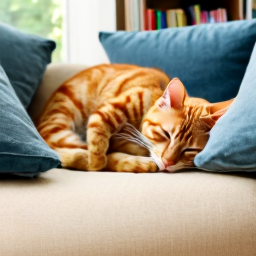}    
  {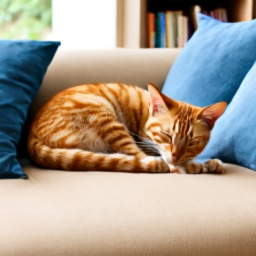} 
  {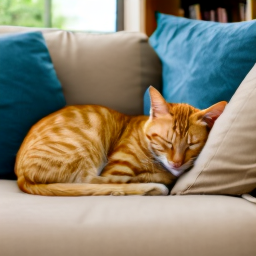}
  {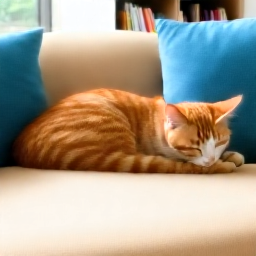}      
  {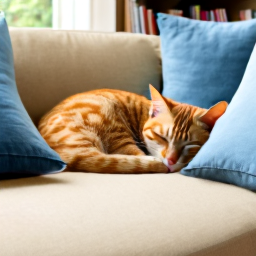}
  {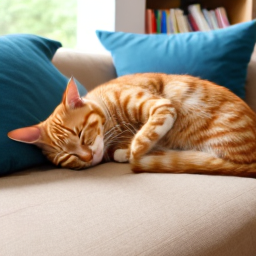}

%\PromptPanelSeven
%  {All the figurines on the bookshelf are white.}
%  {example_image/baseline/00921_sample_0.png} 
%  {example_image/uniform/00921_sample_0.png}    
%  {example_image/static/00921_sample_0.png} 
%  {example_image/fusedit/00921_sample_0.png}      
%  {example_image/time-wise/00921_sample_0.png}
%  {example_image/depth-wise/00921_sample_0.png}
%  {example_image/joint/00921_sample_0.png}

\PromptPanelSeven
  {Three ceramic cups sit to the right of a wooden fork.}
  {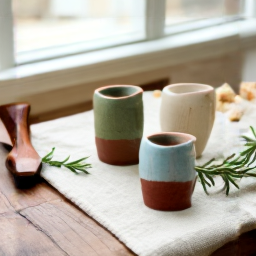} 
  {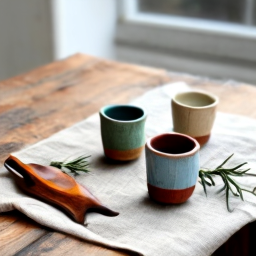}    
  {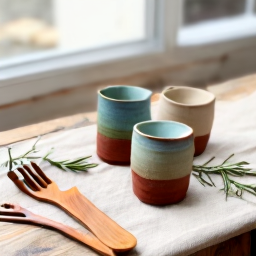} 
  {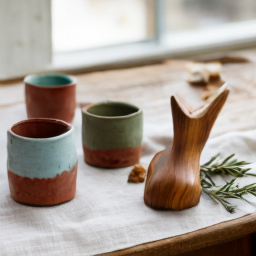}      
  {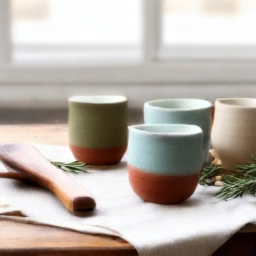}
  {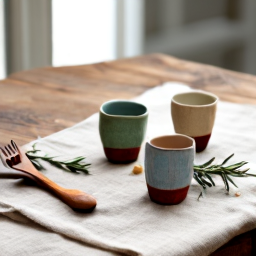}
  {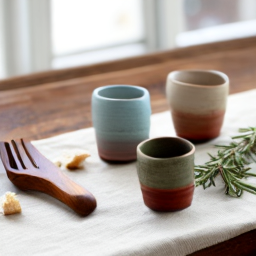}

\PromptPanelSeven
  {Three puppies are standing by the pool and the one in the middle looks more excited than the other two puppies.}
  {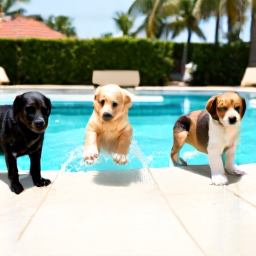} 
  {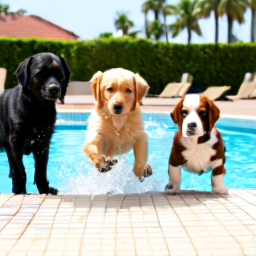}    
  {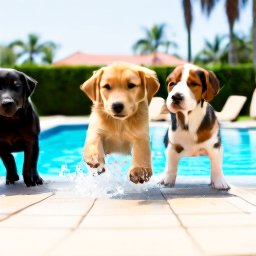} 
  {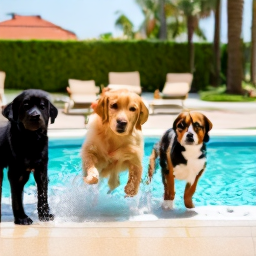}      
  {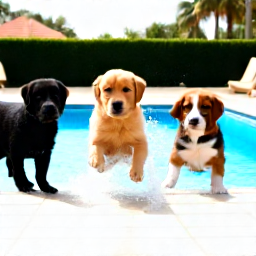}
  {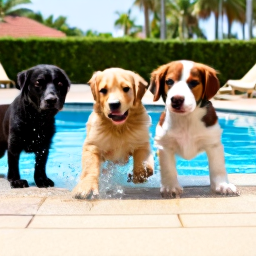}
  {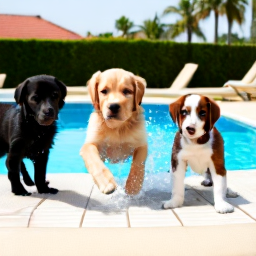}

\end{center}

% Appendix: Fusion weight visualization (more comprehensive comparison)
% Requires: \usepackage{graphicx}
% Recommended: \usepackage{subcaption}

\section{Additional Visualizations of Fusion Weight}
\label{app:weight_viz}

In this section, we provide more detailed visualizations of the fusion weights (App. ~\ref{app:detail_weights}) and an analysis of their variation trends (App. ~\ref{app:trends_analysis}).

\subsection{Detailed Visualization of Fusion Weights}

\begin{figure*}[!htb]
  \centering
  % Row 1
  \begin{subfigure}[t]{0.32\textwidth}
    \centering
    \includegraphics[width=\linewidth]{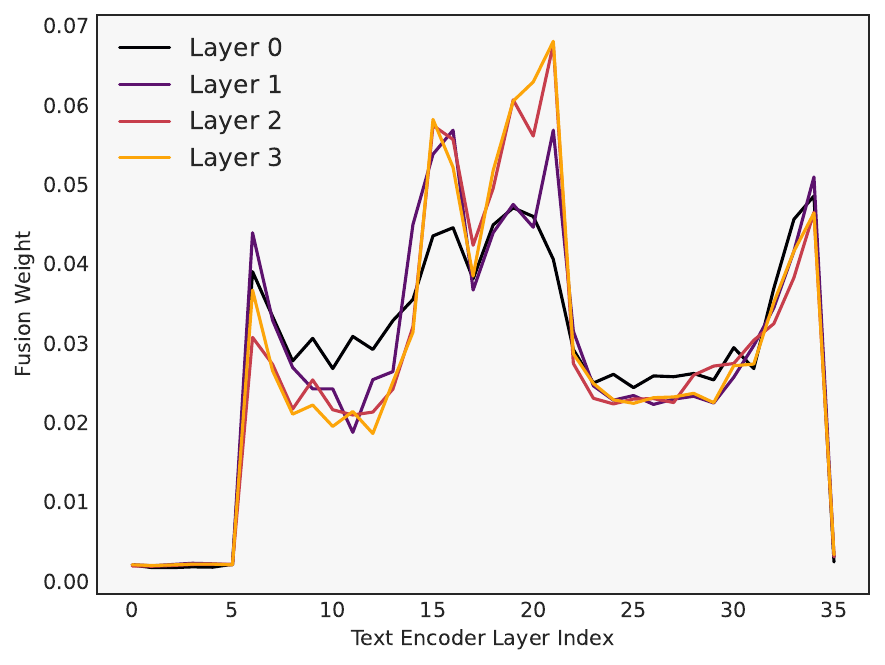}
  \end{subfigure}\hfill
  \begin{subfigure}[t]{0.32\textwidth}
    \centering
    \includegraphics[width=\linewidth]{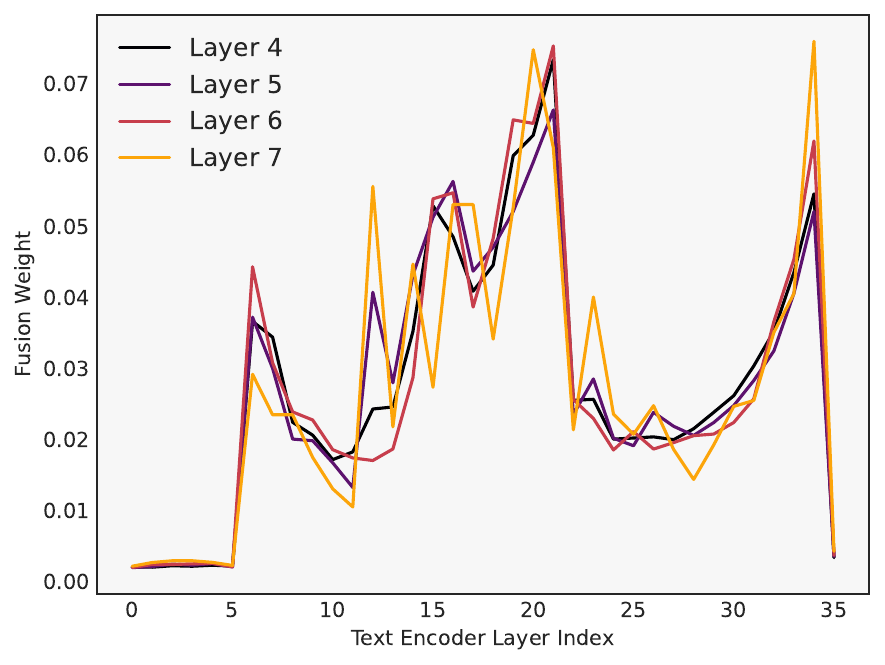}
  \end{subfigure}\hfill
  \begin{subfigure}[t]{0.32\textwidth}
    \centering
    \includegraphics[width=\linewidth]{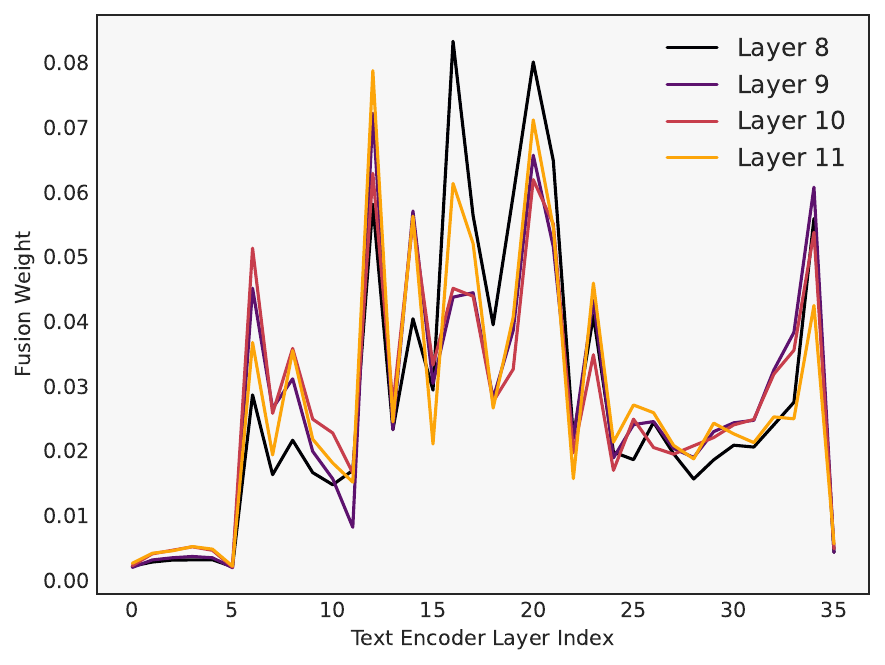}
  \end{subfigure}

  % Row 2
  \begin{subfigure}[t]{0.32\textwidth}
    \centering
    \includegraphics[width=\linewidth]{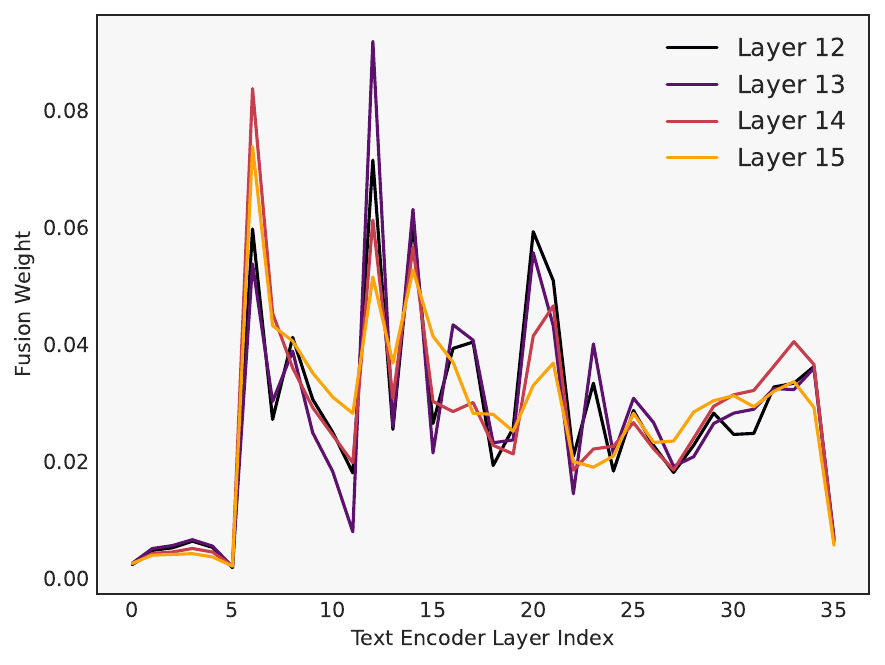}
  \end{subfigure}\hfill
  \begin{subfigure}[t]{0.32\textwidth}
    \centering
    \includegraphics[width=\linewidth]{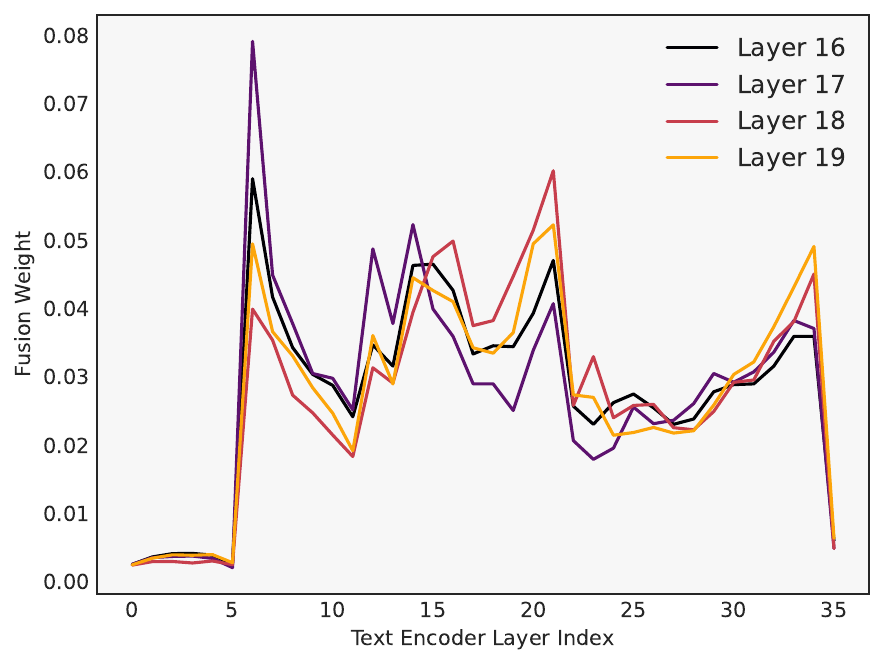}
  \end{subfigure}\hfill
  \begin{subfigure}[t]{0.32\textwidth}
    \centering
    \includegraphics[width=\linewidth]{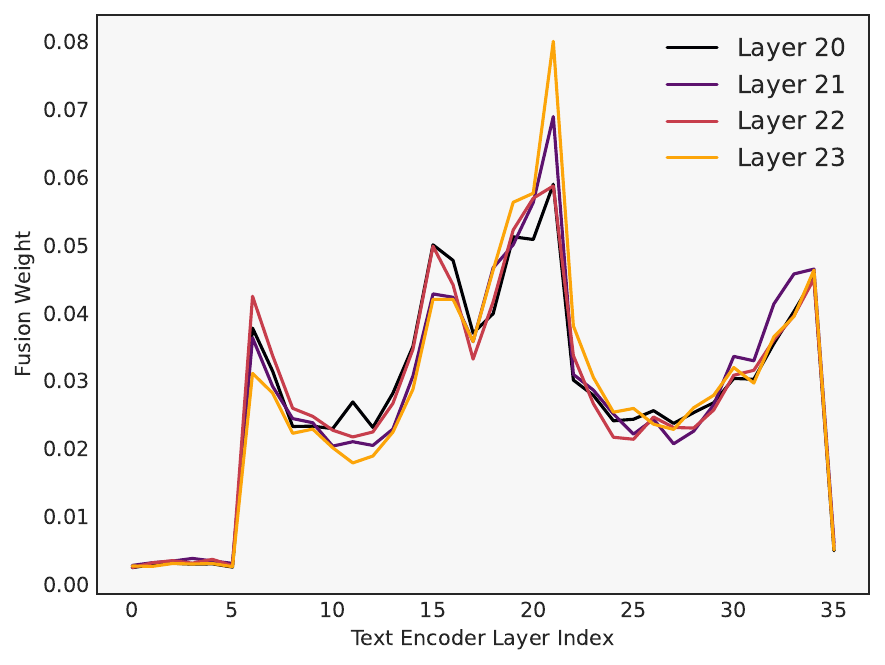}
  \end{subfigure}

  \caption{Depth-wise fusion weights across layers.}
  \label{fig:app_depth_weights}
\end{figure*}

\begin{figure*}[!htb]
  \centering

  \begin{subfigure}[t]{\textwidth}
    \centering
    \includegraphics[width=\linewidth]{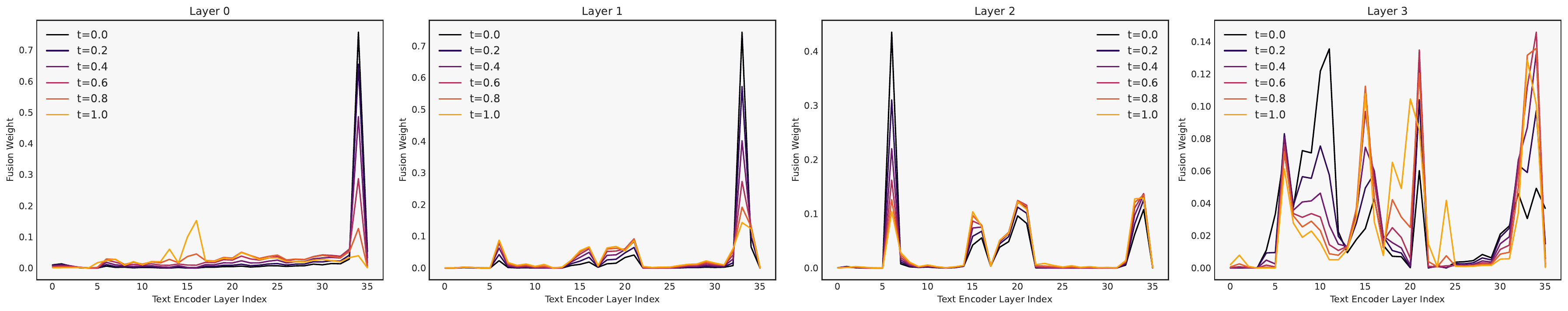}
  \end{subfigure}

  \begin{subfigure}[t]{\textwidth}
    \centering
    \includegraphics[width=\linewidth]{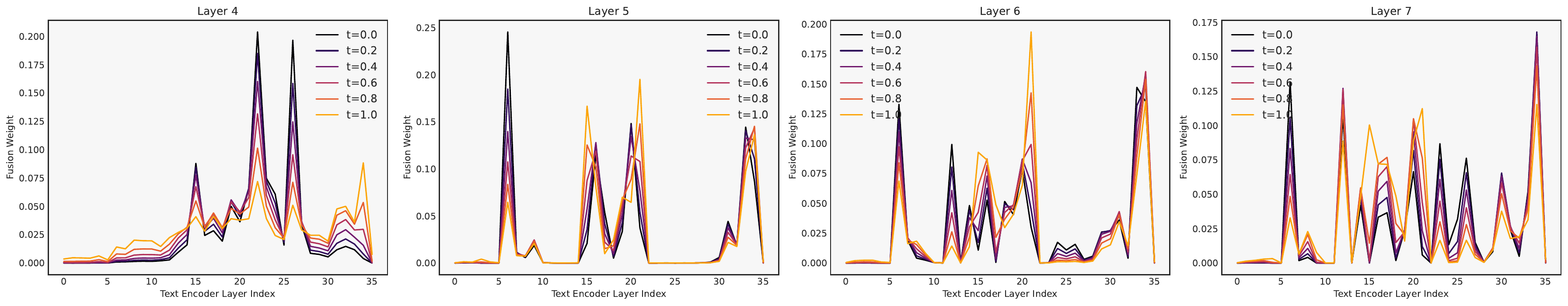}
  \end{subfigure}

  \begin{subfigure}[t]{\textwidth}
    \centering
    \includegraphics[width=\linewidth]{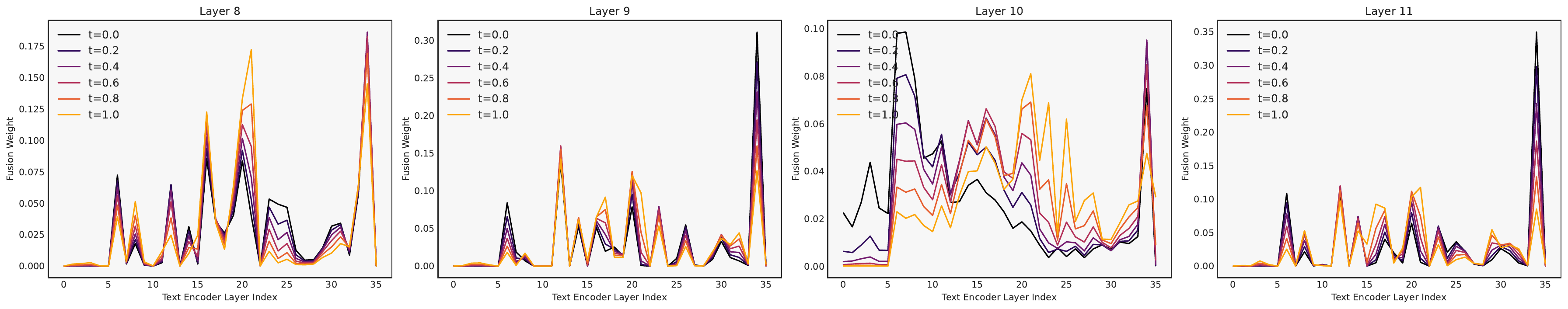}
  \end{subfigure}

  \begin{subfigure}[t]{\textwidth}
    \centering
    \includegraphics[width=\linewidth]{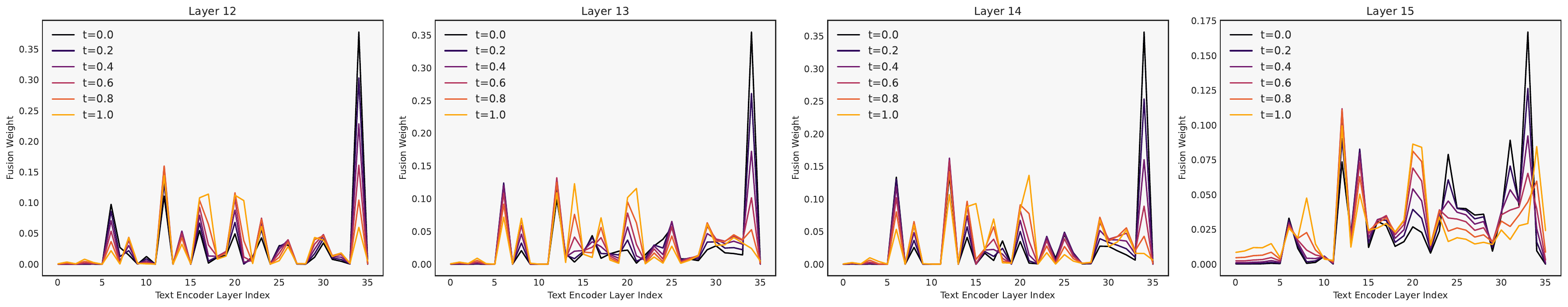}
  \end{subfigure}

  \begin{subfigure}[t]{\textwidth}
    \centering
    \includegraphics[width=\linewidth]{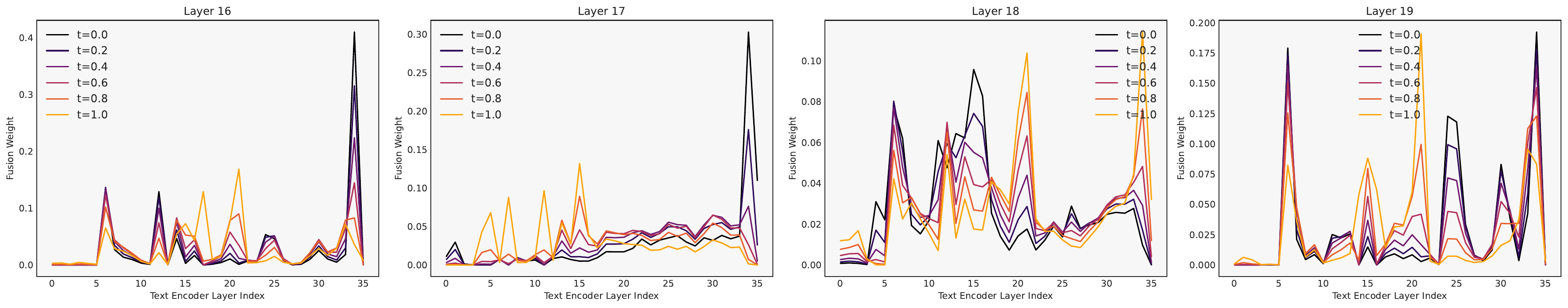}
  \end{subfigure}

  \begin{subfigure}[t]{\textwidth}
    \centering
    \includegraphics[width=\linewidth]{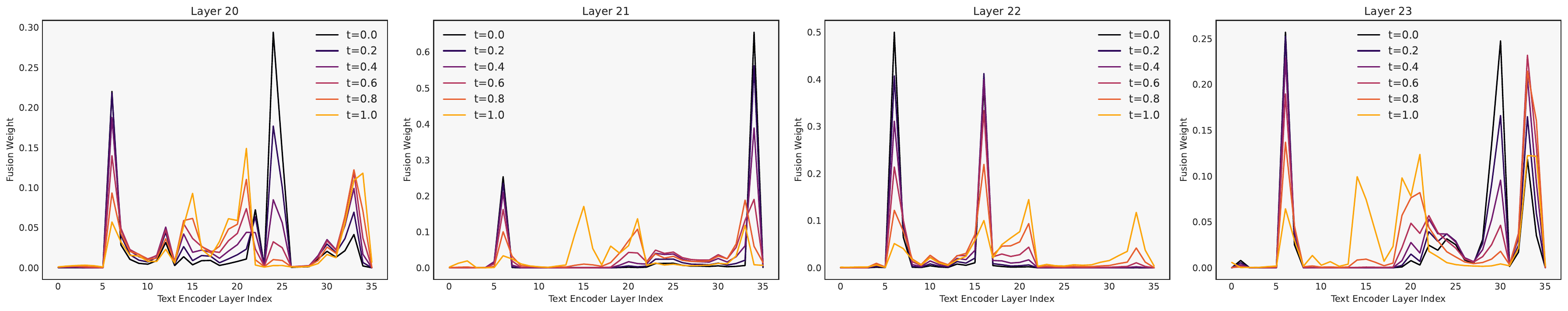}
  \end{subfigure}

  \caption{Joint fusion weights (depth $\times$ time) across layers.}
  \label{fig:app_joint_weights}
\end{figure*}

Figures ~\ref{fig:app_depth_weights} and ~\ref{fig:app_joint_weights} visualize the fusion-weight distributions of the text encoder across different layers and diffusion timesteps under the  depth-wise and joint strategies, respectively.

\label{app:detail_weights}
% ---------------- Figure 1: Depth-wise (6 subfigures, 2 rows) ----------------

% ---------------- Figure 2: Joint (6 subfigures, 1 column) ----------------

\subsection{Weight Evolution and Trend Analysis}
\label{app:trends_analysis}

\begin{figure*}[!htb]
  \centering

  % Row 1: Joint (placed in one row)
  \begin{subfigure}[t]{0.48\textwidth}
    \centering
    \includegraphics[width=\linewidth]{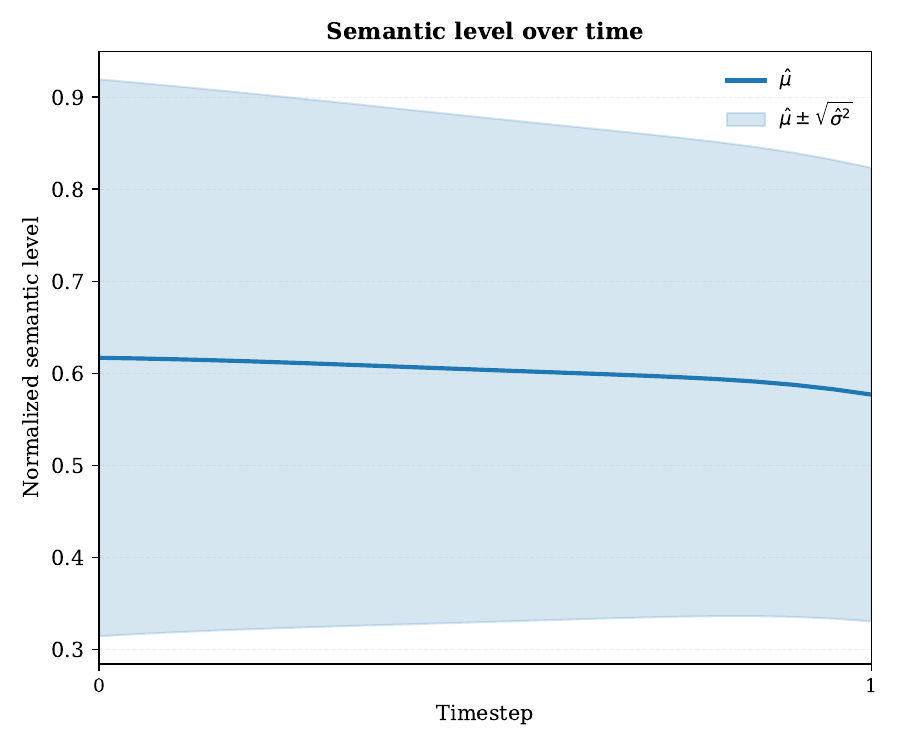}
    \caption{Joint fusion: mean and variance computed on the time-marginal weight distribution.}
    \label{fig:joint_time_marginal_stats}
  \end{subfigure}\hfill
  \begin{subfigure}[t]{0.48\textwidth}
    \centering
    \includegraphics[width=\linewidth]{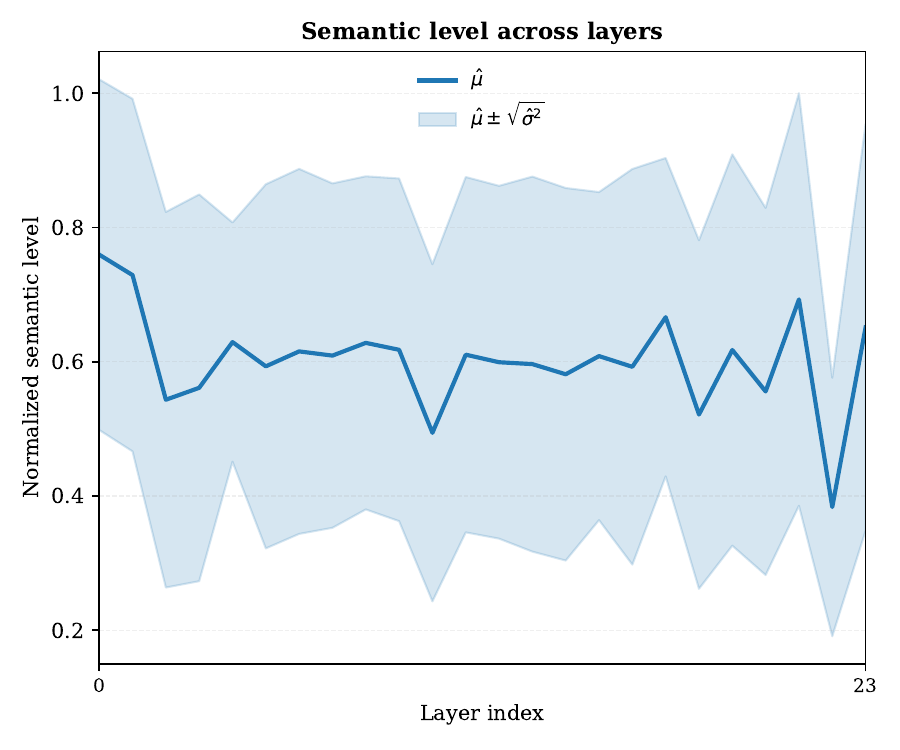}
    \caption{Joint fusion: mean and variance computed on the depth-marginal weight distribution.}
    \label{fig:joint_depth_marginal_stats}
  \end{subfigure}

  % Row 2: Time-wise and Depth-wise
  \begin{subfigure}[t]{0.48\textwidth}
    \centering
    \includegraphics[width=\linewidth]{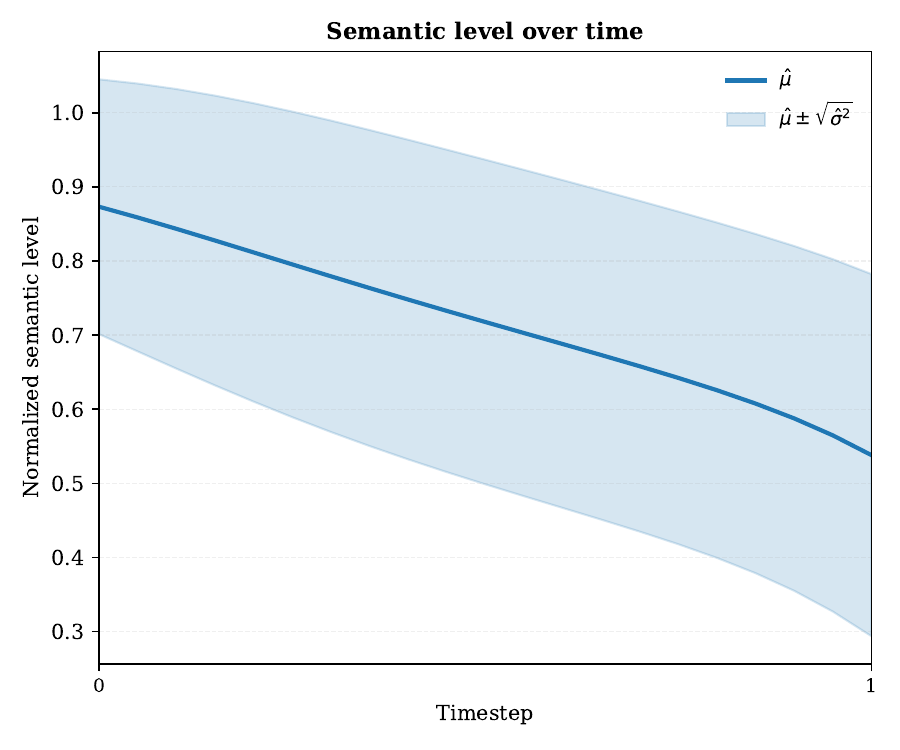}
    \caption{Time-wise fusion: mean and variance trends of the fusion weights across diffusion timesteps.}
    \label{fig:timewise_stats}
  \end{subfigure}\hfill
  \begin{subfigure}[t]{0.48\textwidth}
    \centering
    \includegraphics[width=\linewidth]{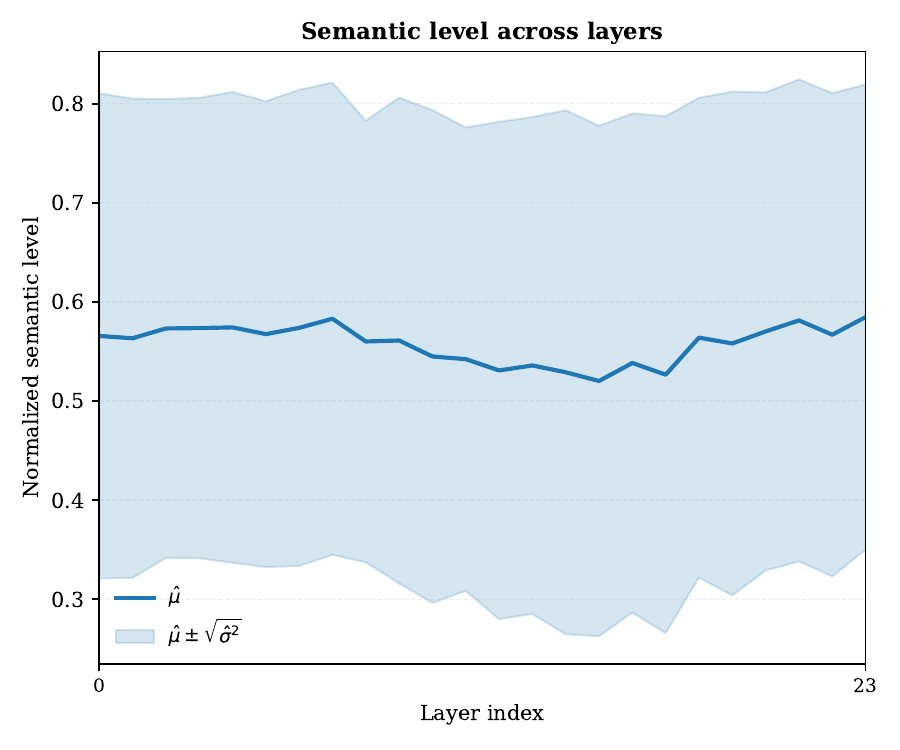}
    \caption{Depth-wise fusion: mean and variance trends of the fusion weights across text-encoder layers.}
    \label{fig:depthwise_stats}
  \end{subfigure}

  \caption{Weight-trend statistics under different fusion strategies. Top row: joint fusion with marginalization over timesteps (left) and depths (right). Bottom row: the corresponding statistics under time-wise (left) and depth-wise (right) fusion.}
  \label{fig:weight_trend_four_panels}
\end{figure*}

To quantify how the fusion weights over text-encoder layers evolve under different diffusion timesteps and depth settings, we treat each fusion-weight vector as a discrete probability distribution supported on uniformly spaced points over $[0,1]$. We then compute its mean and variance as summary statistics, and plot their trends in Figure ~\ref{fig:weight_trend_four_panels}.

Given a fusion-weight vector $\mathbf{w}=\{w_i\}_{i=0}^{L-1}$ of length $L$, we first normalize it as
\begin{equation}
p_i = \frac{w_i}{\sum_{j=0}^{L-1} w_j + \epsilon},
\label{eq:weight_norm}
\end{equation}
where $\epsilon$ is a small positive constant for numerical stability.

We define the uniformly spaced support locations by
\begin{equation}
l_i =
\begin{cases}
\frac{i}{L-1}, & L>1,\\
0, & L=1,
\end{cases}
\qquad i\in\{0,1,\dots,L-1\}.
\label{eq:support_points}
\end{equation}

The mean (semantic center) is defined as
\begin{equation}
\hat{\mu} = \sum_{i=0}^{L-1} p_i\, l_i,
\label{eq:semantic_center}
\end{equation}
and the variance (semantic dispersion) is defined as
\begin{equation}
\hat{\sigma}^2 = \sum_{i=0}^{L-1} p_i\, (l_i - \hat{\mu})^2.
\label{eq:semantic_dispersion}
\end{equation}

Intuitively, this interpretation views fusion weights as a distribution along a ``semantic hierarchy'' axis: $\hat{\mu}$ indicates where the mass concentrates on the axis (favoring earlier vs.\ later layers), while $\hat{\sigma}^2$ measures how dispersed the weights are (larger values imply a less concentrated, more spread-out allocation).

To better understand how fusion weights are allocated across diffusion timesteps and encoder depth, we visualize the statistics in Figure ~\ref{fig:weight_trend_four_panels}. Under the joint strategy, each sample yields a 2D weight map $w(t,l)$ over timestep $t$ and text-encoder layer $l$. To obtain interpretable 1D trends, we form marginal distributions by normalizing along the complementary axis and compute the semantic center $\hat{\mu}$ and dispersion $\hat{\sigma}^2$ on each marginal. 

The results suggest that joint fusion maintains a largely stable semantic center over time: $\hat{\mu}$ stays at a moderately deep-layer regime and only shifts slightly toward shallower layers at later timesteps, while the consistently wide band indicates that the weights do not collapse onto a few layers but preserve broad multi-layer coverage throughout sampling. In contrast, time-wise fusion exhibits pronounced monotonic drift: $\hat{\mu}$ decreases substantially as timesteps progress, indicating a systematic shift from deeper-layer emphasis toward shallower-layer emphasis, which reflects a more time-dependent and less stable allocation of semantic levels. Depth-wise fusion yields smoother curves with smaller fluctuations, suggesting a more consistent and controllable allocation pattern across layers. Overall, joint fusion preserves broad layer coverage while substantially mitigating the deep-to-shallow drift observed in purely time-wise fusion, providing a statistical explanation for its more stable generation behavior.

% Requires:
% \usepackage{graphicx}
% \usepackage{subcaption}

\begin{figure}[t]
  \centering
  \begin{subfigure}[t]{0.48\linewidth}
    \centering
    \includegraphics[width=\linewidth]{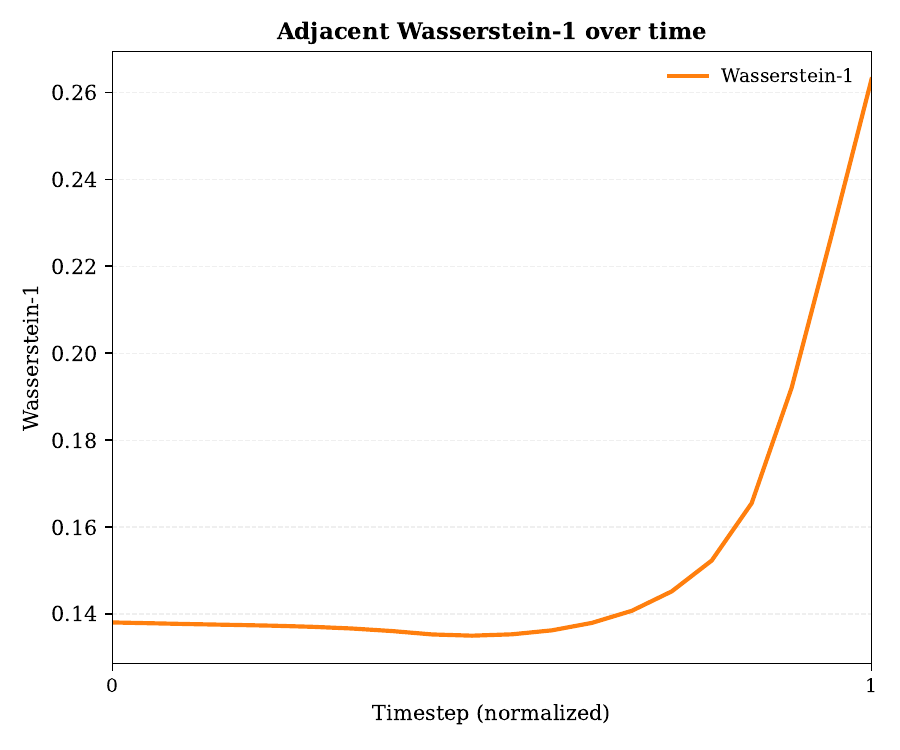}
    \caption{Time-wise fusion: 1-Wasserstein distance between consecutive timesteps.}
    \label{fig:w1_timewise}
  \end{subfigure}\hfill
  \begin{subfigure}[t]{0.48\linewidth}
    \centering
    \includegraphics[width=\linewidth]{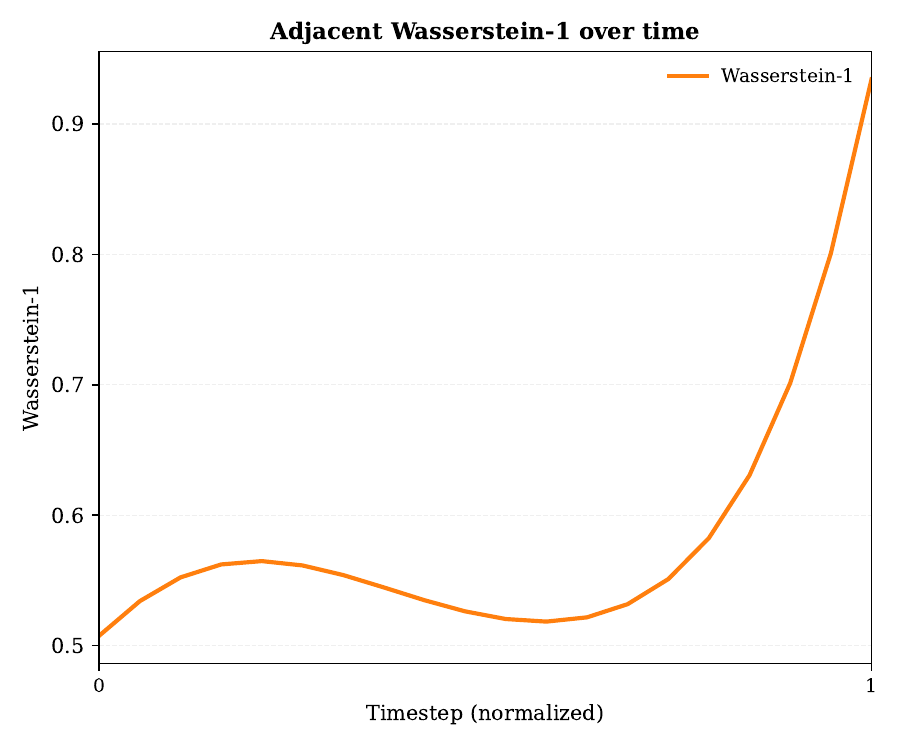}
    \caption{Joint fusion (time-marginal): 1-Wasserstein distance between consecutive timesteps.}
    \label{fig:w1_joint}
  \end{subfigure}
  \caption{Temporal variation speed of fusion weights, measured by the 1-Wasserstein distance between weights at timesteps spaced by 0.05. For joint fusion, we compute the distance on the time-marginal distributions.}
  \label{fig:w1_comparison}
\end{figure}

To compare how fast the fusion weights evolve over time under the time-wise and joint strategies, and to further complement the analysis in Section 5.2, we quantify the discrepancy between weight distributions at consecutive timesteps. Concretely, for each sampled timestep t, we treat its corresponding weight vector as a distribution and compute the 1-Wasserstein distance to the distribution at the previous timestep t-1. For the joint strategy, since the weights are defined over both timesteps and layers, we first marginalize over the layer dimension to obtain a 1D marginal distribution for each timestep, and then compute the 1-Wasserstein distance between consecutive timesteps. We sample 21 timesteps in total, and report the results in Figure ~\ref{fig:w1_comparison}.

As shown in Figure ~\ref{fig:w1_comparison}, the joint strategy yields consistently smaller 1-Wasserstein distances between consecutive timesteps than the time-wise strategy, indicating a substantially smoother temporal evolution of the weight allocation. In light of the train–inference timestep-trajectory mismatch discussed in Section ~\ref{subsec:misalignment}, a simple interpretation is as follows: under a time-wise-only scheme, rapid weight changes across adjacent timesteps can amplify the mismatch-induced deviation, making the semantic conditioning less stable during inference and thus more prone to blur; joint fusion markedly slows down such temporal drift, leading to more coherent conditioning along the inference trajectory and mitigating the resulting degradation.

% Requires:
% \usepackage{graphicx}
% \usepackage{subcaption}

\end{document}